%% file: main.tex
\documentclass[10pt,twocolumn,letterpaper]{article}

\usepackage[pagenumbers]{cvpr} 

\input{preamble}

\definecolor{cvprblue}{rgb}{0.21,0.49,0.74}
\usepackage[pagebackref,breaklinks,colorlinks,citecolor=cvprblue]{hyperref}

\usepackage{graphicx}
\usepackage{subcaption}
\usepackage{makecell}
\usepackage{booktabs}
\usepackage{bm}
\usepackage{xcolor}
\usepackage{float}
\usepackage{multirow}
\usepackage{longtable}
\usepackage{framed}
\usepackage{enumitem}
\usepackage{amsthm,amsmath,amssymb}
\usepackage{pythonhighlight}
\usepackage{appendix}
\usepackage[utf8]{inputenc}
\usepackage{bbding}
\usepackage{tabularx}
\usepackage{dsfont}

\DeclareMathAlphabet{\mathpzc}{OT1}{pzc}{m}{it}

\usepackage{mathtools}
\usepackage{multirow}
\usepackage[noend]{algpseudocode}
\usepackage{color,soul}

\makeatletter
\DeclareRobustCommand\onedot{\futurelet\@let@token\@onedot}
\def\@onedot{\ifx\@let@token.\else.\null\fi\xspace}
\def\eg{\emph{e.g}\onedot} 
\def\ie{\emph{i.e}\onedot} 
 
\def\etc{\emph{etc}\onedot}

\makeatother

\definecolor{blue_}{RGB}{76, 114, 176}
\definecolor{orange_}{RGB}{221, 132, 82}
\definecolor{upload}{RGB}{47, 85, 151}
\definecolor{download}{RGB}{241, 13, 208}
\definecolor{red_}{RGB}{255, 0, 0}
\definecolor{gray_}{RGB}{127, 127, 127}
\definecolor{green_}{RGB}{1, 128, 0}
\definecolor{sjtured_}{RGB}{192, 0, 0}
\definecolor{sjtugreen_}{RGB}{84, 130, 53}
\definecolor{yellow_}{RGB}{255, 192, 0}

\usepackage{pifont}

\usepackage[capitalize]{cleveref}
\crefname{section}{Sec.}{Secs.}
\Crefname{section}{Section}{Sections}
\Crefname{table}{Table}{Tables}
\crefname{table}{Tab.}{Tabs.}

\usepackage{tikz}
\newcommand{\cir}[1]{\tikz[baseline]{%
    \node[anchor=base, draw, circle, inner sep=1pt, scale=0.8]{#1};}}

\usepackage{xspace}
\newcommand{\method}{FedKTL\xspace}

\newcommand{\rulesep}{\unskip\ \vrule\ }

\def\paperID{5850} 
\def\confName{CVPR}
\def\confYear{2024}

\title{An Upload-Efficient Scheme for Transferring Knowledge From a Server-Side Pre-trained Generator to Clients in Heterogeneous Federated Learning}

\author{
    Jianqing Zhang\textsuperscript{\rm 1}\thanks{Work done during internship at AIR.},
    Yang Liu\textsuperscript{\rm 2,3}\thanks{Corresponding authors.},
    Yang Hua\textsuperscript{\rm 4},
    Jian Cao\textsuperscript{\rm 1,5}$^\dagger$ \\
    {\normalsize \textsuperscript{\rm 1}Shanghai Jiao Tong University
    \textsuperscript{\rm 2}Institute for AI Industry Research (AIR), Tsinghua University} \\
    {\normalsize \textsuperscript{\rm 3}Shanghai Artificial Intelligence Laboratory
    \textsuperscript{\rm 4}Queen's University Belfast} \\
    {\normalsize \textsuperscript{\rm 5}Shanghai Key Laboratory of Trusted Data Circulation and Governance in Web3} \\
    {\tt \small tsingz@sjtu.edu.cn, liuy03@air.tsinghua.edu.cn, y.hua@qub.ac.uk, cao-jian@sjtu.edu.cn}
}

\begin{document}
\maketitle

\begin{abstract}
    Heterogeneous Federated Learning (HtFL) enables \underline{task-specific knowledge} sharing among clients with different model architectures while preserving privacy. Despite recent research progress, transferring knowledge in HtFL is still difficult due to data and model heterogeneity. To tackle this, we introduce a public pre-trained generator (e.g., StyleGAN or Stable Diffusion) as the bridge and propose a new upload-efficient knowledge transfer scheme called Federated Knowledge-Transfer-Loop (\textbf{\method}). It can produce task-related prototypical image-vector pairs via the generator's inference on the server. With these pairs, each client can transfer \underline{common knowledge} from the generator to its local model through an additional supervised local task. We conduct extensive experiments on four datasets under two types of data heterogeneity with 14 heterogeneous models, including CNNs and ViTs. Results show that our \method surpasses seven state-of-the-art methods by up to \textbf{\textcolor{green_}{7.31\%}}. Moreover, our knowledge transfer scheme is applicable in cloud-edge scenarios with only one edge client. Code: \url{https://github.com/TsingZ0/FedKTL}
\end{abstract}

\section{Introduction}

Recently, there has been a growing trend for companies to develop custom models tailored to their specific needs~\cite{brown2020language, jiao2020tinybert, sun2022survey, hu2023survey, feng2023ernie}. However, the problem of insufficient data has persistently plagued model training in specific fields, such as medicine~\cite{ponzio2019dealing, ayan2018data, candemir2021training}. Federated Learning (FL) is a popular approach to tackle this problem by training models collaboratively among multiple clients (\eg, companies or edge devices) while preserving privacy on clients~\cite{kairouz2019advances, li2020federated}.  Traditional FL (tFL) focuses on training a global model for all clients and is unable to fulfill clients' personalized needs due to data heterogeneity among clients~\cite{karimireddy2020scaffold, MLSYS2020_38af8613}. Consequently, personalized FL (pFL) has emerged as a solution to train customized models for each client~\cite{li2021ditto, zhang2022fedala, zhang2023gpfl, yang2023dynamic}. 

However, most pFL methods still assume homogeneous client models~\cite{li2021ditto, zhang2022fedala, zhang2023gpfl}, which may not adequately cater to the specific needs of companies and individuals~\cite{yi2023fedgh}. Besides, as the size of the model increases, both tFL and pFL incur significant communication costs when transmitting model parameters~\cite{zhuang2023foundation}. Furthermore, exposing clients’ model parameters also raises privacy and intellectual property (IP) concerns~\cite{zhang2024fedtgp, li2021survey, zhang2018protecting, wang2023model}. Recently, Heterogeneous Federated Learning (HtFL) frameworks have been proposed to consider both data and model heterogeneity~\cite{yi2023fedgh, tan2022fedproto}. It explores novel knowledge-sharing schemes that go beyond sharing the entire client models. 

Most existing HtFL methods adopt knowledge distillation (KD) techniques~\cite{hinton2015distilling} and design various knowledge-sharing frameworks based on a global dataset~\cite{lin2020ensemble, zhang2021parameterized}, a global auxiliary model~\cite{wu2022communication, zhang2022fine}, or global class-wise prototypes~\cite{zhang2024fedtgp, tan2022fedproto, tan2022federated}. 
However, global datasets' availability and quality as well as their \textit{relevance to clients' tasks} significantly impact the effectiveness of KD~\cite{zhang2023towards}. Directly replacing the global dataset with a pre-trained generator has a minimal impact since most generators are pre-trained to generate unlabeled data \textit{within the domain of their pre-training data}~\cite{karras2019style, karras2020analyzing}. As for the global auxiliary model, it introduces a substantial communication overhead due to the need to transmit it in each communication iteration. Although sharing class-wise prototypes is communication-efficient, they can only carry limited global knowledge to clients, which is insufficient for clients’ model training needs. Furthermore, the prototypes extracted by heterogeneous models are biased, hindering the attainment of uniformly separated global prototypes on the server~\cite{zhang2024fedtgp}. 

Thus, we propose an upload-efficient knowledge transfer scheme called \textit{Federated Knowledge-Transfer-Loop (\textbf{\method})}, which takes advantage of the compactness of prototypes and the pre-existing knowledge from a server-side public pre-trained generator. 
\method can (1) use the generator on the server to produce a handful of global prototypical image-vector pairs tailored to clients' tasks, and (2) transfer pre-existing common knowledge from the generator to each client model via an additional \textit{supervised} local task using these image-vector pairs. 
We develop \method by addressing the following three questions. \textbf{Q1}: \textit{How to upload unbiased prototypes while maintaining upload efficiency?} \textbf{Q2 (the core challenge)}: \textit{How to adapt any given pre-trained generator to clients' tasks without fine-tuning it?} \textbf{Q3}: \textit{How to transfer the generator's knowledge to client models regardless of the semantics of the generated images?} 

\begin{figure}[t]
  \begin{subfigure}{0.24\linewidth}
    \includegraphics[width=\linewidth]{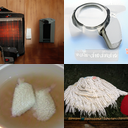}
    \caption{Valid vecs}\label{subfig:wspace}
  \end{subfigure}\hfill
  \begin{subfigure}{0.24\linewidth}
    \includegraphics[width=\linewidth]{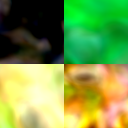}
    \caption{Random vecs}\label{subfig:random}
  \end{subfigure}\hfill
  \begin{subfigure}{0.24\linewidth}
    \includegraphics[width=\linewidth]{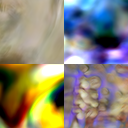}
    \caption{Prototypes}\label{subfig:proto}
  \end{subfigure}\hfill
  \begin{subfigure}{0.24\linewidth}
    \includegraphics[width=\linewidth]{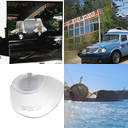}
    \caption{Aligned vecs}\label{subfig:our}
  \end{subfigure}
  \caption{The images ($64\times 64$) generated by StyleGAN-XL~\cite{sauer2022StyleGAN} with different kinds of inputs. ``vecs'' is short for vectors. }
  \label{fig:intro}
  \vspace{-6pt}
\end{figure}

For \textbf{Q1}, inspired by FedETF~\cite{Li_2023_ICCV}, we replace each client's classifier with an ETF (equiangular tight frame) classifier~\cite{yang2022inducing, Li_2023_ICCV} to let clients generate unbiased prototypes. Then, we upload these unbiased prototypes to the server for efficiency. For \textbf{Q2}, we align the domain formed by prototypes with the generator's inherent valid latent domain to generate informative images, as \textit{these two domains are not naturally aligned}. As shown in \cref{fig:intro}, the generator can generate clear images given valid vectors. 
However, it tends to generate blurry and uninformative images given invalid latent vectors (such as random vectors or prototypes). To generate prototype-induced clear images, we propose a \textit{lightweight trainable feature transformer} on the server to convert prototypes to aligned vectors within the valid input domain, while preserving the class-wise discrimination relevant to clients' classification tasks. 
For \textbf{Q3}, we first aggregate aligned vectors for each class to obtain latent centroids and generate corresponding images to form image-vector pairs. Then we conduct an additional supervised local task to only enhance the client model's feature extraction ability using these pairs, thereby reducing the semantic relevance requirements between the generated images and local data. 

We evaluate our \method via extensive experiments on four datasets with two types of data heterogeneity and 14 model architectures using a StyleGAN~\cite{karras2019style, karras2020analyzing, karras2021alias, sauer2022StyleGAN} or a Stable Diffusion~\cite{rombach2022high} on the server. Our \method can outperform seven state-of-the-art methods by at most \textbf{\textcolor{green_}{7.31\%}} in accuracy. We also show that \method is upload-efficient and one prototypical image-vector pair per class is sufficient for knowledge transfer, which only demands minimal inference of the generator on the server in each iteration.

\section{Related Work}

\subsection{Heterogeneous Federated Learning (HtFL)}

HtFL offers the advantage of preserving both privacy and model IP while catering to personalized model architecture requirements~\cite{tan2022fedproto, yi2023fedgh, fang2023robust}. In terms of the level of model heterogeneity, we classify existing HtFL methods into three categories: group heterogeneity, partial heterogeneity, and full heterogeneity. 

\textit{Group-heterogeneity-based HtFL methods} distribute multiple groups of homogeneous models to clients, considering their diverse communication and computing capabilities~\cite{lin2020ensemble, diao2020heterofl}. They typically form groups by sampling submodels from a server model~\cite{diao2020heterofl, horvath2021fjord, wen2022federated}. In this paper, we do not consider this kind of model heterogeneity due to IP protection concerns and client customization limitations. 

\textit{Partial-heterogeneity-based HtFL methods}, \eg, LG-FedAvg~\cite{liang2020think}, FedGen~\cite{zhu2021data}, and FedGH~\cite{yi2023fedgh}, allow the main parts of the clients' models to be heterogeneous but assume the remaining (small) parts to be homogeneous. However, clients can only access limited global knowledge through the small global part. Despite training a global representation generator, FedGen primarily utilizes it to introduce global knowledge for the small classifier rather than the remaining main part (\ie, the feature extractor). Therefore, the data insufficiency problem still exists for the main part. 

\textit{Full-heterogeneity-based HtFL methods} do not impose restrictions on the architectures of client models. Classic KD-based HtFL approaches ~\cite{li2019fedmd, yu2022multimodal} share model outputs on a global dataset. However, obtaining such a dataset can be difficult in practice~\cite{zhang2023towards}. Instead of relying on a global dataset, FML~\cite{shen2020federated} and FedKD~\cite{wu2022communication} utilize mutual distillation~\cite{zhang2018deep} between a small auxiliary global model and client models. However, in the early iterations when both the auxiliary model and client models have poor performance, there is a risk of transferring uninformative knowledge between each other~\cite{li2023smkd}. Another approach is to share class prototypes, like FedDistill~\cite{jeong2018communication}, FedProto~\cite{tan2022fedproto}, and FedPCL~\cite{tan2022federated}. 
However, the phenomenon of classifier bias has been extensively observed in FL when dealing with heterogeneous data~\cite{luo2021no, Li_2023_ICCV}. The bias becomes more pronounced when both the models and the data exhibit heterogeneity, leading to biased prototypes, thereby posing challenges in aggregating class-wise global knowledge~\cite{zhang2024fedtgp}.

\subsection{ETF Classifier}

When training a model on balanced data reaches its terminal stage, the neural collapse~\cite{papyan2020prevalence} phenomenon occurs. In this phenomenon, prototypes and the classifier vectors converge to form a simplex ETF, where the vectors are normalized, and the pairwise angles between them are maximized and identical (balanced). Since a simplex ETF represents an ideal classifier, some centralized methods~\cite{yang2022inducing, yang2022neural} propose generating a random simplex ETF matrix to replace the original classifier and guiding the feature extractor training using the fixed ETF classifier in imbalanced scenarios. To address the data heterogeneity issue in FL, FedETF~\cite{Li_2023_ICCV} also proposes to replace the original classifier for each client with a fixed ETF classifier. However, FedETF assumes the presence of homogeneous models and follows FedAvg to transfer global knowledge. 
Inspired by these methods, we utilize the ETF classifier to enable heterogeneous client models to generate unbiased prototypes and facilitate class-wise global knowledge aggregation on the server. 

\section{Method}

\subsection{Preliminaries}

Several concepts in various generators, such as StyleGAN~\cite{karras2019style} and Stable Diffusion~\cite{rombach2022high}, share similarities when generating contents, despite potential differences in their nomenclature. Without loss of generality, we primarily focus on introducing the generator components based on StyleGAN's architecture here for convenience. Most existing StyleGANs contain two components: a mapping network $G_m$ and a synthesis network $G_s$. The space formed by the latent vectors between $G_m$ and $G_s$ is called ``$\mathcal{W}$ space''. 

\begin{figure}[ht]
	\centering
	\includegraphics[width=\linewidth]{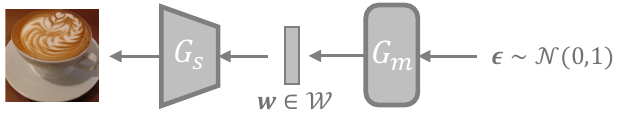}
	\caption{An illustration of the generating process (from right to left) when utilizing StyleGAN-XL as an example. The solid border of $G_s$ and $G_m$ means ``with frozen parameters''. }
	\label{fig:StyleGAN}
\end{figure}

In \cref{fig:StyleGAN}, we show an example of the StyleGAN-XL~\cite{sauer2022StyleGAN} employed in our \method. Given a vector ${\bm \epsilon}$ (typically a normally distributed noise vector) as the input, it transforms ${\bm \epsilon}$ to a latent vector ${\bm w} \in \mathbb{R}^H$ through $G_m$, \ie, ${\bm w} = G_m({\bm \epsilon}) \in \mathcal{W}$. Then, it generates an image $I$ by further transforming ${\bm w}$ with $G_s$, \ie, $I = G_s({\bm w})$. ${\bm w}$ is the only factor that controls the content of $I$. While the valid vectors in $\mathcal{W}$ can produce clear and informative images, not all vectors in $\mathbb{R}^H$ are valid and possess the same capability. 

\subsection{Problem Statement}

In HtFL, one server and $N$ clients collaborate to train client models for a multi-classification task of $C$ classes. Client $i$ owns private data $\mathcal{D}_i$ and builds its model $g_i$ (parameterized by ${\bm W}_i$) with a customized architecture. 
Formally, the objective is 
$\min_{\{{\bm W}_i\}_{i=1}^N} \sum_{i=1}^{N} \frac{n_i}{n} L_i({\bm W}_i, \mathcal{D}_i)$, 
where $n_i = |\mathcal{D}_i|$, $n = \sum_{i=1}^{N} n_i$, and $L_i$ is the local loss function. 

\subsection{Our \method}

\begin{figure*}[t]
  \begin{subfigure}{0.74\linewidth}
    \includegraphics[width=\linewidth]{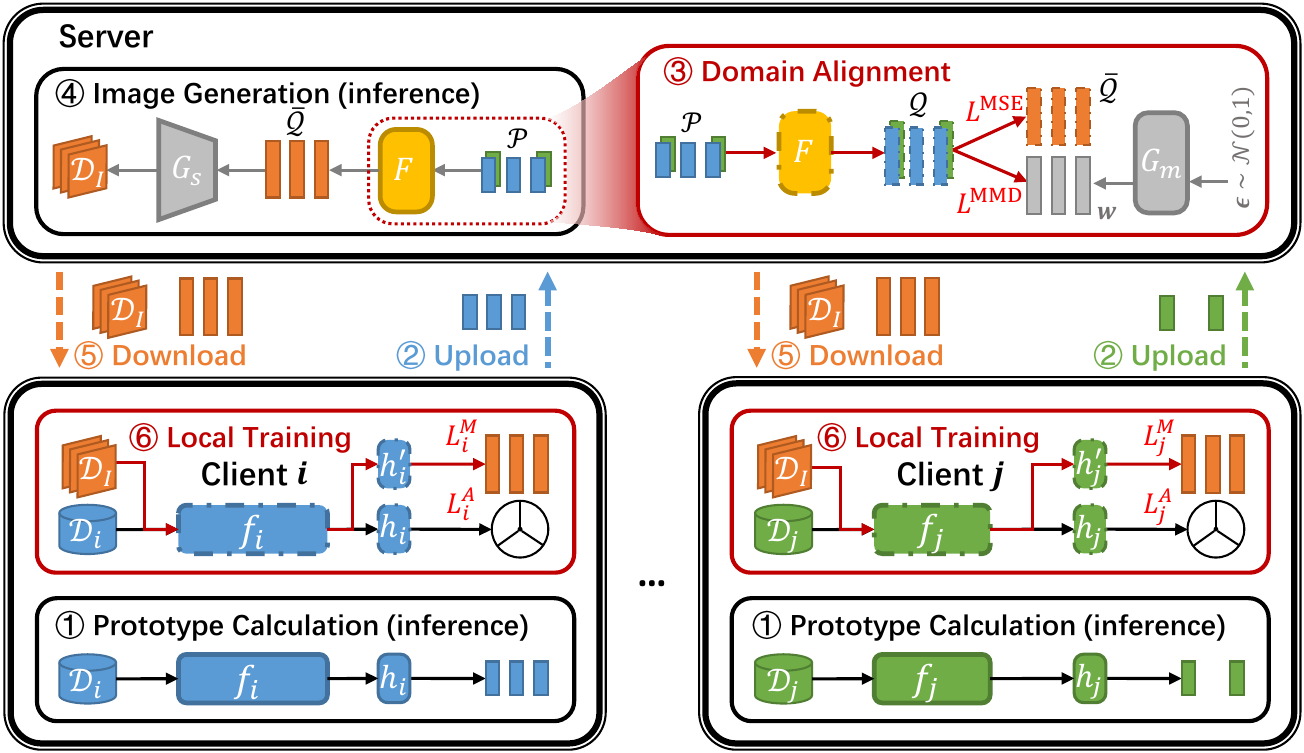}
    \caption{The framework of our \method in one communication iteration for HtFL.}\label{subfig:main}
  \end{subfigure}\hfill
  \begin{minipage}[t]{0.25\linewidth}
    \vspace{-77mm}
    \begin{subfigure}{\linewidth}
      \includegraphics[width=\linewidth]{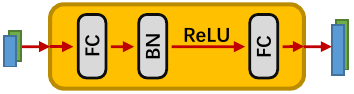}
      \caption{The feature transformer ($F$). }\label{subfig:F}
      \vspace{3mm}
    \end{subfigure}
    \begin{subfigure}{\linewidth}
      \includegraphics[width=\linewidth]{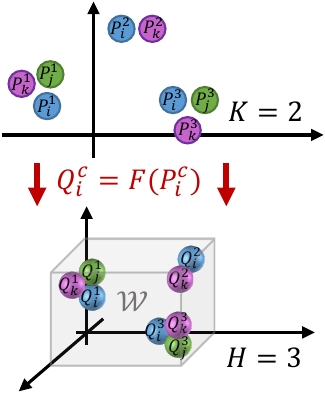}
      \caption{A domain alignment example.}\label{subfig:example}
    \end{subfigure}
  \end{minipage}
  \caption{An example of our \method for a 3-class classification task. (a) Rounded and slender rectangles denote models and representations, respectively; dash-dotted and solid borders denote updating and frozen components, respectively; the segmented circle represents the ETF classifier. (b) The feature transformer ($F$) contains two FC layers and one Batch Normalization~\cite{ioffe2015batch} (BN) layer. (c) An example of the domain alignment step with $K=2$ and $H=3$; one cluster represents one class. \textit{Best viewed in color.} }
  \label{fig:all}
\end{figure*}

\subsubsection{Overview}
In \cref{subfig:main}, we illustrate six key steps of the knowledge-transfer-loop in our proposed \method framework.
\textbf{\cir{1}} After local training, each client generates class-wise prototypes. \textbf{\cir{2}} Each client uploads prototypes to the server. \textbf{\cir{3}} The server trains a feature transformer (denoted by $F$ with parameter ${\bm W}_F$) to transform and align client prototypes to latent vectors. \textbf{\cir{4}} With the trained $F$, the server first obtains the class-wise latent centroid $\bar{\mathcal{Q}}$, which is the averaged latent vectors within each class, and then generates images $\mathcal{D}_I$ by inputting $\bar{\mathcal{Q}}$ into $G_s$. \textbf{\cir{5}} Each client downloads the prototypical image-vector pairs $\{\mathcal{D}_I, \bar{\mathcal{Q}}\}$ from the server. \textbf{\cir{6}} Each client locally trains $g_i$ and $h'_i$ using $\mathcal{D}_i$, $\mathcal{D}_I$, and $\bar{\mathcal{Q}}$, where $h'_i$ is an additional linear projection layer (parameterized by ${\bm W}_{h'_i}$) used to change the dimension of feature representations. Notice that $|\bar{\mathcal{Q}}| = |\mathcal{D}_I| = C \ll |\mathcal{D}_i|$. 

\subsubsection{ETF Classifier and Prototype Generation}

The local loss $L_i$ consists of two components: $L_i^A$, which is the loss corresponding to $\mathcal{D}_i$, and $L_i^M$, which is the loss for knowledge transfer using $\mathcal{D}_I$ and $\bar{\mathcal{Q}}$. For clarity, we only describe $L_i^A$ here and leave the details of $L_i^M$ to \cref{sec:lmse}. 

To address the biased prototype issue, inspired by FedETF~\cite{Li_2023_ICCV}, we replace the original classifiers of given model architectures with identical ETF classifiers and add a linear projection layer (one Fully Connected (FC) layer) $h_i$ to the feature extractor $f_i$. In this way, we encourage each local model $g_i$ to generate unbiased prototypes that are aligned with the globally identical ETF classifier vectors. 
$f_i$ and $h_i$ have parameters ${\bm W}_{f_i}$ and ${\bm W}_{h_i}$, respectively. Thus, we have $g_i = h_i \circ f_i$ and ${\bm W}_i = \{{\bm W}_{f_i}, {\bm W}_{h_i}\}$. 

Specifically, we first synthesize a simplex ETF ${\bm V} = [{\bm v}_1, \ldots, {\bm v}_C]$, where ${\bm V} = \sqrt{\frac{C}{C-1}}{\bm U}({\bm I}_C - \frac{1}{C}{\bm 1}_C{\bm 1}_C^T) \in \mathbb{R}^{K\times C}$ and the dimension of the ETF space $K \ge C-1$. $\forall c \in [C], {\bm v}_c \in \mathbb{R}^{K}$ and the $L_2$-norm $||{\bm v}_c||_2 = 1$. ${\bm U}$ allows a rotation, ${\bm U} \in \mathbb{R}^{K\times C}$, ${\bm U}^T{\bm U} = {\bm I}_C$, ${\bm I}_C$ is an identity matrix, and ${\bm 1}_C$ is a vector with all ones. Besides, $\forall c_1, c_2 \in [C]$ and $c_1 \ne c_2$, we have $\cos{\theta} = -\frac{1}{C-1}$, where $\theta$ is the angle between ${\bm v}_{c_1}$ and ${\bm v}_{c_2}$. Furthermore, $\theta$ is also the maximum angle to equally separate $C$ vectors~\cite{Li_2023_ICCV, papyan2020prevalence, yang2022inducing}. 
Then, we distribute ${\bm V}$ to all clients. 

Next, for a given input ${\bm x}$ on client $i$, we compute logits by measuring the cosine similarity~\cite{nguyen2010cosine} between $g_i({\bm x})$ and each vector in ${\bm V}$. As the ArcFace loss~\cite{deng2019arcface} is popular for enhancing supervised learning when using cosine similarity for classification, we apply it during local training: 
\begin{equation}
    L_i^A = \mathbb{E}_{({\bm x}, y) \sim \mathcal{D}_i} - \log{\frac{e^{s(\cos{(\theta_{y} + m)})}}{e^{s(\cos{(\theta_{y} + m)})} + \sum_{c=1, c \ne y}^C e^{s\cos{\theta_c}}}}, \label{eq:arcface}
\end{equation}
where $\theta_{y}$ is the angle between $g_i({\bm x})$ and ${\bm v}_{y}$, $s$ and $m$ are the re-scale and additive hyperparameters~\cite{deng2019arcface}, respectively. 

After local training, we fix $g_i$ and collect prototypes $\mathcal{P}_i = \{{\bm P}_i^c\}_{c \in \mathcal{C}_i}$ in the ETF space, where $\mathcal{C}_i$ is a set of class labels on client $i$. Formally, ${\bm P}_i^c = \mathbb{E}_{({\bm x}, c) \sim \mathcal{D}_i^c} \ g_i({\bm x})\in \mathbb{R}^K$, where $\mathcal{D}_i^c$ refers to the subset of $\mathcal{D}_i$ containing data points belonging to class $c$. Uploading $\mathcal{P}_i$ to the server only requires $|\mathcal{C}_i| \times K$ elements to communicate, where $|\mathcal{C}_i| \le C$. 

\subsubsection{Domain Alignment and Image Generation}

For simplicity, we assume full client participation here, although \method supports partial participation. 
With clients' prototypes $\mathcal{P} = \{{\bm P}_i^c\}_{i\in [N], c \in \mathcal{C}_i}$ on the server, we devise a trainable feature transformer $F$ (see \cref{subfig:F}) to convert $\mathcal{P}$ into valid latent vectors $\mathcal{Q} = \{{\bm Q}_i^c\}_{i\in [N], c \in \mathcal{C}_i}$, where ${\bm Q}_i^c = F({\bm P}_i^c) \in \mathbb{R}^H$, in $\mathcal{W}$ space. To maintain $\mathcal{Q}$'s relationship with clients' classification tasks, we first preserve $\mathcal{Q}$'s class-wise discrimination by training $F$ with
\vspace{-2pt}
\begin{equation}
    L^{\rm MSE} = \frac{1}{C} \sum_{c=1}^{C} \frac{1}{|\mathcal{M}_c|} \sum_{i \in \mathcal{M}_c} \ell(F({\bm P}_i^c), {\bm Q}^c), \label{eq:smse}
\vspace{-1pt}
\end{equation}
where $\mathcal{M}_c$ is the client set owning class $c$, the global class-wise centroid ${\bm Q}^c = \frac{1}{|\mathcal{M}_c|} \sum_{j \in \mathcal{M}_c} F({\bm P}_j^c)$, and $\ell$ is the Mean Squared Error (MSE)~\cite{tuchler2002minimum} between two vectors. 
Then, we use the Maximum Mean Discrepancy (MMD) loss~\cite{li2021fedphp} to align the domain formed by $\mathcal{Q}$ with the valid input domain of $G_s$ in $\mathcal{W}$: 
\begin{equation}
    L^{\rm MMD} = ||\mathbb{E}_{{\bm Q} \sim \mathcal{Q}} \ \phi({\bm Q}) - \mathbb{E}_{{\bm w} \sim \mathcal{W}} \ \phi({\bm w})||_{\mathcal{H}}^2. 
\end{equation}
${\bm w}$ is randomly sampled using $G_m$, $\phi$ is a feature map induced by a kernel function $\kappa$, \ie, $\kappa({\bm a}, {\bm b}) = \langle \phi({\bm a}), \phi({\bm b}) \rangle$, and $\mathcal{H}$ is a reproducing kernel Hilbert space~\cite{long2015learning, li2021fedphp}. We combine these two losses to form the server loss $L = L^{\rm MMD} + \lambda L^{\rm MSE}$, where $\lambda$ is a hyper-parameter. 
We show a domain alignment example in \cref{subfig:example}. 

After training $F$ on the server, we generate one image per class by inputting global centroids $\bar{\mathcal{Q}} = \{{\bm Q}^c\}_{c=1}^C$ into $G_s$, so only $C$ times of inference for $G_s$ is required in each iteration. Formally, we generate $\mathcal{D}_I = \{I^c\}_{c=1}^C$, where $I^c = G_s({\bm Q}^c)$, and distribute paired class-wise $\mathcal{D}_I$ and $\bar{\mathcal{Q}}$ to clients for additional local supervised learning. 

\subsubsection{Transferring Pre-existing Global Knowledge}
\label{sec:lmse}

Then, client $i$ conducts local training with the integrated local loss $L_i = L_i^A + \mu L_i^M$, where $\mu$ is a hyper-parameter. $L_i^M$ is the additional supervised task to transfer pre-existing knowledge from the generator and inject common and shared information into the feature extractor. Formally, 
\vspace{-5pt}
\begin{equation}
    L_i^M = \frac{1}{C} \sum_{c=1}^C \ell(h'_i(f_i(I^c)), {\bm Q}^c), \label{eq:L_i^M}
\end{equation}
where $h'_i$ is a linear projection layer that outputs vectors with dimension $H$. Since $\mathcal{D}_I$ and $\bar{\mathcal{Q}}$ are the output-input pairs of $G_s$ and serve as the input-output pairs for $h'_i \circ f_i$, we can transfer common knowledge from $G_s$ to $h'_i \circ f_i$. Since $h'_i$ is mainly used for dimension transformation rather than knowledge learning, we initialize ${\bm W}_{h'_i}$ in an identical way for all clients in each iteration, which does not introduce additional communication costs. This approach minimizes the biased knowledge acquired by $h'_i$ and facilitates the transfer of common knowledge from $G_s$ to $f_i$.

\subsubsection{Privacy-Preserving Discussion}

Our \method preserves privacy in three folds. (1) We introduce an identical ETF classifier for all clients to generate unbiased prototypes, which contain little private information. (2) 
The generated images belong to the generator's inherent output domain, so they are much different from the client's local data (see \cref{fig:generated}). (3) We keep all the model parameters locally on clients without sharing. \textit{See the Appendix for further analysis and experimental results. }

\section{Experiments}

\begin{table*}[h]
  \centering
  \resizebox{\linewidth}{!}{
    \begin{tabular}{l|*{4}{c}|*{4}{c}}
    \toprule
    Settings & \multicolumn{4}{c|}{Pathological Setting} & \multicolumn{4}{c}{Practical Setting} \\
    \midrule
    Datasets & Cifar10 & Cifar100 & Flowers102 & Tiny-ImageNet & Cifar10 & Cifar100 & Flowers102 & Tiny-ImageNet \\
    \midrule
    LG-FedAvg & 86.82$\pm$0.26 & 57.01$\pm$0.66 & 58.88$\pm$0.28 & 32.04$\pm$0.17 & 84.55$\pm$0.51 & 40.65$\pm$0.07 & 45.93$\pm$0.48 & 24.06$\pm$0.10 \\
    FedGen & 82.83$\pm$0.65 & 58.26$\pm$0.36 & 59.90$\pm$0.15 & 29.80$\pm$1.11 & 82.55$\pm$0.49 & 38.73$\pm$0.14 & 45.30$\pm$0.17 & 19.60$\pm$0.08 \\
    FedGH & 86.59$\pm$0.23 & 57.19$\pm$0.20 & 59.27$\pm$0.33 & 32.55$\pm$0.37 & 84.43$\pm$0.31 & 40.99$\pm$0.51 & 46.13$\pm$0.17 & 24.01$\pm$0.11 \\
    FML & 87.06$\pm$0.24 & 55.15$\pm$0.14 & 57.79$\pm$0.31 & 31.38$\pm$0.15 & 85.88$\pm$0.08 & 39.86$\pm$0.25 & 46.08$\pm$0.53 & 24.25$\pm$0.14 \\
    FedKD & 87.32$\pm$0.31 & 56.56$\pm$0.27 & 54.82$\pm$0.35 & 32.64$\pm$0.36 & 86.45$\pm$0.10 & 40.56$\pm$0.31 & 48.52$\pm$0.28 & 25.51$\pm$0.35 \\
    FedDistill & 87.24$\pm$0.06 & 56.99$\pm$0.27 & 58.51$\pm$0.34 & 31.49$\pm$0.38 & 86.01$\pm$0.31 & 41.54$\pm$0.08 & 49.13$\pm$0.85 & 24.87$\pm$0.31 \\
    FedProto & 83.39$\pm$0.15 & 53.59$\pm$0.29 & 55.13$\pm$0.17 & 29.28$\pm$0.36 & 82.07$\pm$1.64 & 36.34$\pm$0.28 & 41.21$\pm$0.22 & 19.01$\pm$0.10 \\
    \midrule
    \method & \textbf{88.43$\pm$0.13} & \textbf{62.01$\pm$0.28} & \textbf{64.72$\pm$0.62} & \textbf{34.74$\pm$0.17} & \textbf{87.63$\pm$0.07} & \textbf{46.94$\pm$0.23} & \textbf{53.16$\pm$0.08} & \textbf{28.17$\pm$0.18} \\
    \bottomrule
    \end{tabular}}
  \caption{The test accuracy (\%) on four datasets in the pathological and practical settings using HtFE$_8$.}
    \label{tab:datasets_acc}
\end{table*}

\subsection{Setup}

\noindent\textbf{Datasets and baseline methods. } In this paper, we evaluate our \method on four image datasets, \ie, Cifar10~\cite{krizhevsky2009learning}, Cifar100~\cite{krizhevsky2009learning}, Tiny-ImageNet~\cite{chrabaszcz2017downsampled}, and Flowers102~\cite{nilsback2008automated} (8K images with 102 classes). Besides, we compare \method with seven state-of-the-art HtFL methods, including LG-FedAvg~\cite{liang2020think}, FedGen~\cite{zhu2021data}, FedGH~\cite{yi2023fedgh}, FML~\cite{shen2020federated}, FedKD~\cite{wu2022communication}, FedDistill~\cite{jeong2018communication}, and FedProto~\cite{tan2022fedproto}. 

\noindent\textbf{Model heterogeneity scenarios. } LG-FedAvg, FedGen, and FedGH assume the classifier to be homogeneous. Unless explicitly specified, we consider model heterogeneity for the main model part, \ie, using Heterogeneous Feature Extractors (HtFE), for a fair comparison. Specifically, we denote the model heterogeneity scenarios by ``HtFE$_X$'', where the suffix number $X$ represents the degree of model heterogeneity, and we utilize a total of $X$ model architectures in HtFL. The larger the $X$ is, the more heterogeneous the scenario is. Given $N$ clients, we distribute the $(i \mod X)$th model architecture to client $i, i\in [N]$ and reinitialize its parameters. For instance, we use HtFE$_8$ by default, which includes eight model architectures: 4-layer CNN~\cite{mcmahan2017communication}, GoogleNet~\cite{szegedy2015going}, MobileNet\_v2~\cite{sandler2018mobilenetv2}, ResNet18, ResNet34, ResNet50, ResNet101, and ResNet152~\cite{he2016deep}. The model architectures in HtFE$_8$ cover both small and large models. The feature dimensions $K'$ before classifiers are different in these model architectures, which cannot meet the assumptions of FedGH, FedKD, and FedProto, so we add an average pooling layer~\cite{szegedy2015going} before classifiers and set $K'=512$ by default for all model architectures. 

\noindent\textbf{Data heterogeneity. } Following prior arts~\cite{mcmahan2017communication, zhu2021data, zhang2023gpfl} in the FL field, we consider two data heterogeneity scenarios, including the pathological setting~\cite{tan2022fedproto, Zhang2023fedcp, zhang2024fedtgp} and the practical setting~\cite{tan2022federated, zhang2022fedala, zhang2023eliminating}. In the pathological setting, following FedALA~\cite{zhang2022fedala}, we assign unbalanced data of 2/10/10/20 classes to each client from a total of 10/100/102/200 classes from Cifar10/Cifar100/Flowers102/Tiny-ImageNet datasets without overlap. As for the practical setting, following GPFL~\cite{zhang2023gpfl}, we assign a proportion $q_{c, i}$ of data from a subset that contains all the data belonging to class $c$ in a public dataset to client $i$, where $q_{c, i} \sim Dir(\beta)$, $Dir(\beta)$ is Dirichlet distribution and $\beta$ is typically set to 0.1~\cite{lin2020ensemble}. 

\noindent\textbf{General Implementation Details. } We combine the above model and data heterogeneity to simulate HtFL scenarios. Besides, we split the local data into a training set and a test set with a ratio of 3:1 following~\cite{zhang2022fedala, Zhang2023fedcp}. The performance of clients' models is assessed using their respective test sets, and these results (\eg, test accuracy) are then averaged to gauge the performance of an HtFL method. Following FedAvg, we set the client batch size to 10 and run one training epoch with SGD~\cite{zhang2015deep}, \ie, $\lfloor \frac{n_i}{10} \rfloor$ SGD steps, on the client in each iteration. Besides, we set the client learning rate $\eta_c=0.01$ and the total communication iterations to 1000. We run three trials and report the mean and standard deviation of the numerical results. We simulate HtFL scenarios on 20 clients with a client participation ratio $\rho=1$, and we experiment on 50, 100, and 200 clients with $\rho=0.5$. 

\noindent\textbf{Implementation Details for Our \method. } We set $\mu=50$, $\lambda=1$, $K=C$, $\eta_S=0.01$, $B_S=100$, and $E_S=100$ by default on all tasks, where $\eta_S$, $B_S$, and $E_S$ represent the learning rate, batch size, and number of epochs for training $F$ on the server. Besides, we use Adam~\cite{kingma2015adam} for $F$ training following FedGen and set $s=64$ and $m=0.5$ following ArcFace loss~\cite{deng2019arcface}. By default, we use a public pre-trained StyleGAN-XL~\cite{sauer2022StyleGAN} as the server-side generator (not used during clients' inference), which is one of the latest StyleGANs. It has approximately 0.13 billion model parameters and is trained on a large-scale ImageNet dataset~\cite{deng2009imagenet} to generate images with a resolution of $64\times64$. To ensure compatibility with clients' models, we rescale the generated images on the server to match the resolution of clients' data before downloading them. 
\textit{See the Appendix for the experiments using Stable Diffusion or only one edge client. }

\begin{table*}[t]
  \centering
  \resizebox{\linewidth}{!}{
    \begin{tabular}{l|*{5}{c}|*{3}{c}}
    \toprule
    Settings & \multicolumn{5}{c|}{Different Degrees of Model Heterogeneity} & \multicolumn{3}{c}{Large Client Amount ($\rho=0.5$)}\\
    \midrule
    & HtFE$_2$ & HtFE$_3$ & HtFE$_4$ & HtFE$_9$ & HtM$_{10}$ & 50 Clients & 100 Clients & 200 Clients \\
    \midrule
    LG-FedAvg & 46.61$\pm$0.24 & 45.56$\pm$0.37 & 43.91$\pm$0.16 & 42.04$\pm$0.26 & --- & 37.81$\pm$0.12 & 35.14$\pm$0.47 & 27.93$\pm$0.04 \\
    FedGen & 43.92$\pm$0.11 & 43.65$\pm$0.43 & 40.47$\pm$1.09 & 40.28$\pm$0.54 & --- & 37.95$\pm$0.25 & 34.52$\pm$0.31 & 28.01$\pm$0.24 \\
    FedGH & 46.70$\pm$0.35 & 45.24$\pm$0.23 & 43.29$\pm$0.17 & 43.02$\pm$0.86 & --- & 37.30$\pm$0.44 & 34.32$\pm$0.16 & 29.27$\pm$0.39 \\
    FML & 45.94$\pm$0.16 & 43.05$\pm$0.06 & 43.00$\pm$0.08 & 42.41$\pm$0.28 & 39.87$\pm$0.09 & 38.47$\pm$0.14 & 36.09$\pm$0.28 & 30.55$\pm$0.52 \\
    FedKD & 46.33$\pm$0.24 & 43.16$\pm$0.49 & 43.21$\pm$0.37 & 42.15$\pm$0.36 & 40.36$\pm$0.12 & 38.25$\pm$0.41 & 35.62$\pm$0.55 & 31.82$\pm$0.50 \\
    FedDistill & 46.88$\pm$0.13 & 43.53$\pm$0.21 & 43.56$\pm$0.14 & 42.09$\pm$0.20 & 40.95$\pm$0.04 & 38.51$\pm$0.36 & 36.06$\pm$0.24 & 31.26$\pm$0.13 \\
    FedProto & 43.97$\pm$0.18 & 38.14$\pm$0.64 & 34.67$\pm$0.55 & 32.74$\pm$0.82 & 36.06$\pm$0.10 & 33.03$\pm$0.42 & 28.95$\pm$0.51 & 24.28$\pm$0.46 \\
    \midrule
    \method & \textbf{48.06$\pm$0.19} & \textbf{49.83$\pm$0.44} & \textbf{47.06$\pm$0.21} & \textbf{50.33$\pm$0.35} & \textbf{45.84$\pm$0.15} & \textbf{43.16$\pm$0.82} & \textbf{39.73$\pm$0.87} & \textbf{34.24$\pm$0.45} \\
    \bottomrule
    \end{tabular}}
  \caption{The test accuracy (\%) on Cifar100 in the practical setting with different degrees of model heterogeneity or large client amounts. }
    \label{tab:hetero}
\end{table*}

\subsection{Performance Comparison}

We show the test accuracy of all the methods on four datasets in \cref{tab:datasets_acc}, where \method achieves superior performance than baselines in HtFL scenarios. Specifically, our \method outperforms counterparts by up to \textbf{\textcolor{green_}{5.40\%}} in test accuracy on Cifar100 in the practical setting. Besides, our \method demonstrates greater superiority in the practical setting compared to the pathological setting. The number of generated images in $\mathcal{D}_I$ equals the number of classes $C$, so $|\mathcal{D}_I|$ is 10/100/102/200 for Cifar10/Cifar100/Flowers102/Tiny-ImageNet. Even with only 10 images in $\mathcal{D}_I$, our \method can still perform excellently on Cifar10 in two data heterogeneous settings.

\subsection{Impact of Model Heterogeneity}

We further assess \method on the other five scenarios with incremental model heterogeneity. Specifically, we consider HtFE$_2$, HtFE$_3$, HtFE$_4$, HtFE$_9$, and HtM$_{10}$. HtFE$_2$ includes 4-layer CNN and ResNet18. HtFE$_3$ includes ResNet10~\cite{zhong2017deep}, ResNet18, and ResNet34. HtFE$_4$ includes 4-layer CNN, GoogleNet, MobileNet\_v2, and ResNet18. HtFE$_9$ includes ResNet4, ResNet6, and ResNet8~\cite{zhong2017deep}, ResNet10, ResNet18, ResNet34, ResNet50, ResNet101, and ResNet152. HtM$_{10}$ contains all the model architectures in HtFE$_8$ plus another two architectures ViT-B/16~\cite{dosovitskiy2020image} and ViT-B/32~\cite{dosovitskiy2020image}. ``HtM'' is short for heterogeneous models, where classifiers are also heterogeneous. 
LG-FedAvg, FedGen, and FedGH are \textit{not applicable} for HtM$_{10}$ due to the different classifier architectures of ResNets and ViTs. We allocate model architectures in HtM$_{10}$ to clients using the method introduced for HtFE$_X$. We show the test accuracy in \cref{tab:hetero}. For almost all the baselines, their performance deteriorates as model heterogeneity increases, resulting in an accuracy drop of at least \textbf{3.53\%} from HtFE$_2$ to HtFE$_9$. 
In contrast, \method attains its best performance with HtFE$_9$, outperforming baselines by \textbf{\textcolor{green_}{7.31\%}}. 

\subsection{Partial Participation with More Clients}

To study the scalability of our \method in HtFL settings with more clients, we introduce three scenarios with 50, 100, and 200 clients on HtFE$_8$, respectively, by splitting the Cifar100 dataset differently. With 200 participating clients, each class has an average of only eight samples for training. 
We consider partial client participation and set $\rho = 0.5$ in each iteration in these three scenarios. 
Notice that comparing the accuracy between these scenarios is unreasonable because both the number of clients and the amount of client data change when splitting Cifar100 into different numbers of clients' datasets. As shown in \cref{tab:hetero}, our \method maintains its superiority even with a large number of clients and partial client participation. 

\subsection{Impact of Number of Client Training Epochs}

\begin{table}[ht]
  \centering
  \resizebox{\linewidth}{!}{
    \begin{tabular}{l|*{3}{c}}
    \toprule
     & $E=5$ & $E=10$ & $E=20$ \\
    \midrule
    LG-FedAvg & 40.33$\pm$0.15 & 40.46$\pm$0.08 & 40.93$\pm$0.23 \\
    FedGen & 40.00$\pm$0.41 & 39.66$\pm$0.31 & 40.07$\pm$0.12 \\
    FedGH & 41.09$\pm$0.25 & 39.87$\pm$0.27 & 40.22$\pm$0.41 \\
    FML & 39.08$\pm$0.27 & 37.97$\pm$0.19 & 36.02$\pm$0.22 \\
    FedKD & 41.06$\pm$0.13 & 40.36$\pm$0.20 & 39.08$\pm$0.33 \\
    FedDistill & 41.02$\pm$0.30 & 41.29$\pm$0.23 & 41.13$\pm$0.41 \\
    FedProto & 38.04$\pm$0.52 & 38.13$\pm$0.42 & 38.74$\pm$0.51 \\
    \midrule
    \method & \textbf{46.18$\pm$0.34} & \textbf{45.70$\pm$0.27} & \textbf{45.57$\pm$0.23} \\
    \bottomrule
    \end{tabular}}
  \caption{The test accuracy (\%) on Cifar100 in the practical setting using HtFE$_8$ with large $E$.}
    \label{tab:largeE}
\end{table}

\noindent Training more epochs on clients before uploading can save communication resources~\cite{mcmahan2017communication}. Here, we increase the number of client training epochs and study its effects. From \cref{tab:largeE}, we observe that most of the methods, except for FML and FedKD, can maintain their performance even with a large value of $E$. Notably, our \method maintains its superior performance across different values of $E$. Since FML and FedKD learn an auxiliary model following the scheme of FedAvg, the auxiliary model tends to learn more biased information during local training with a larger value of $E$, which may deteriorate the auxiliary model aggregation~\cite{qu2022rethinking}. 

\subsection{Impact of Feature Dimensions}

\begin{table}[ht]
  \centering
  \resizebox{\linewidth}{!}{
    \begin{tabular}{l|*{3}{c}}
    \toprule
     & $K'=64$ & $K'=256$ & $K'=1024$ \\
    \midrule
    LG-FedAvg & 39.69$\pm$0.25 & 40.21$\pm$0.11 & 40.46$\pm$0.01 \\
    FedGen & 39.78$\pm$0.36 & 40.38$\pm$0.36 & 40.83$\pm$0.25 \\
    FedGH & 39.93$\pm$0.45 & 40.80$\pm$0.40 & 40.19$\pm$0.37 \\
    FML & 39.89$\pm$0.34 & 40.95$\pm$0.09 & 40.26$\pm$0.16 \\
    FedKD & 41.06$\pm$0.18 & 41.14$\pm$0.35 & 40.72$\pm$0.25 \\
    FedDistill & 41.69$\pm$0.10 & 41.66$\pm$0.15 & 40.09$\pm$0.27 \\
    FedProto & 30.71$\pm$0.65 & 37.16$\pm$0.42 & 31.21$\pm$0.27 \\
    \midrule
    \method & \textbf{46.46$\pm$0.41} & \textbf{47.81$\pm$0.43} & \textbf{45.91$\pm$0.54} \\
    \bottomrule
    \end{tabular}}
  \caption{The test accuracy (\%) on Cifar100 in the practical setting using HtFE$_8$ with different $K'$.}
\end{table}

\noindent Here, we study the impact of $K'$ on the test accuracy. Most of the methods achieve their best performance when setting $K'=256$, except for the methods that share classifiers, such as LG-FedAvg and FedGen. Using a larger value of $K'$, FedProto can generate prototypes with dimension $K'$ and upload more client information to the server. In contrast, our \method generates prototypes after the projection layer ($h_i, i \in [N]$) with another dimension of $K = C < K'$. This dimension is fixed, \ie, $K=100$, for the 100-classification problem on Cifar100. 

\subsection{Communication Cost}

\begin{table}[ht]
  \centering
  \resizebox{!}{!}{
    \begin{tabular}{l|rr|c}
    \toprule
     & Upload & Download & Accuracy \\
    \midrule
    LG-FedAvg & 1.03M & 1.03M & 40.65$\pm$0.07 \\
    FedGen & 1.03M & 7.66M & 38.73$\pm$0.14 \\
    FedGH & 0.46M & 1.03M & 40.99$\pm$0.51\\
    FML & 18.50M & 18.50M & 39.86$\pm$0.25 \\
    FedKD & 16.52M & 16.52M & 40.56$\pm$0.31 \\
    FedDistill & 0.09M & 0.20M & 41.54$\pm$0.08 \\
    FedProto & 0.46M & 1.02M & 36.34$\pm$0.28 \\
    \midrule
    \method & 0.09M & 7.17M & \textbf{46.94$\pm$0.23} \\
    \bottomrule
    \end{tabular}}
  \caption{The upload and download overhead per iteration using HtFE$_8$ on Cifar100 with 20 clients in the practical setting. ``M'' is short for million. The accuracy column is referred from \cref{tab:datasets_acc}. }
    \label{tab:comm}
\end{table}

Our \method exhibits excellent performance while maintaining an affordable communication cost, as shown in \cref{tab:comm}. Specifically, \method exhibits lower upload and download costs compared to FedGen, FML, and FedKD. Notably, the upload cost of our approach is the lowest among all the baselines, since we set $K=C$ for our \method. Besides, the upload overhead required by \method is much less than the download one, which is suitable for real-world scenarios, where the uplink speed is typically lower than the downlink speed~\cite{li2023fast}. The upload-efficient characteristic of \method highlights its practicality for knowledge transfer in HtFL. 

\subsection{Adapting to Various Pre-Trained StyleGAN3s}

\begin{table}[ht]
  \centering
  \resizebox{\linewidth}{!}{
    \begin{tabular}{l|*{3}{c}}
    \toprule
     & $\lambda=0.05$ & $\lambda=0.1$ & $\lambda=0.5$ \\
    \midrule
    AFHQv2 & 26.82$\pm$0.32 & \textbf{27.05$\pm$0.26} & 26.32$\pm$0.52 \\
    Bench & 27.71$\pm$0.25 & \textbf{28.36$\pm$0.42} & 27.56$\pm$0.50 \\
    FFHQ-U & \textbf{27.28$\pm$0.23} & 27.21$\pm$0.35 & 26.59$\pm$0.47 \\
    WikiArt & 27.37$\pm$0.51 & \textbf{27.48$\pm$0.33} & 27.30$\pm$0.15 \\
    \bottomrule
    \end{tabular}}
  \caption{The test accuracy (\%) on Tiny-ImageNet in the practical setting using HtFE$_8$ with different pre-trained StyleGAN3s, which are represented by the names of the pre-training datasets. }
    \label{tab:generators}
\end{table}

\begin{figure}[ht]
  \begin{subfigure}{0.19\linewidth}
    \includegraphics[width=\linewidth]{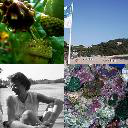}
    \caption{Client \#1}\label{subfig:Client1}
  \end{subfigure}\hfill
  \rulesep
  \begin{subfigure}{0.19\linewidth}
    \includegraphics[width=\linewidth]{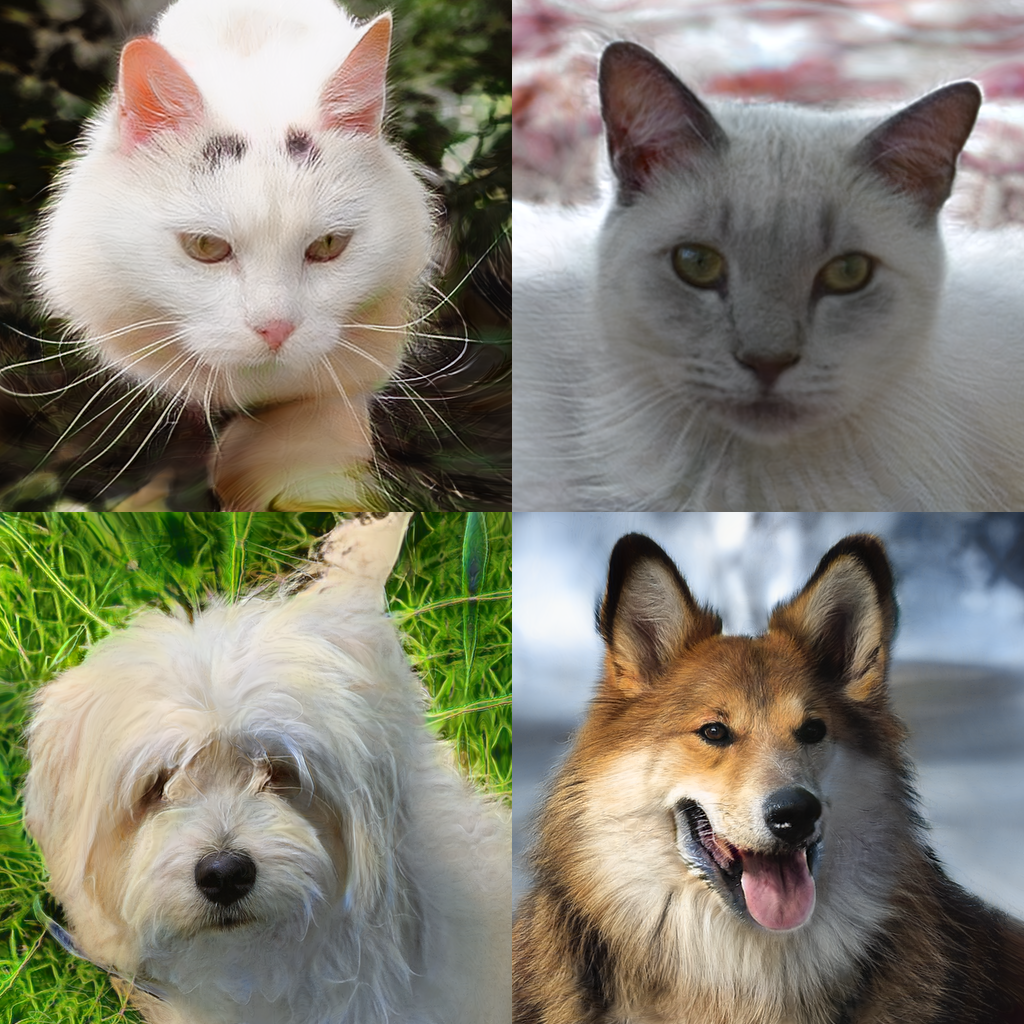}
    \caption{AFHQv2}\label{subfig:AFHQv2}
  \end{subfigure}\hfill
  \begin{subfigure}{0.19\linewidth}
    \includegraphics[width=\linewidth]{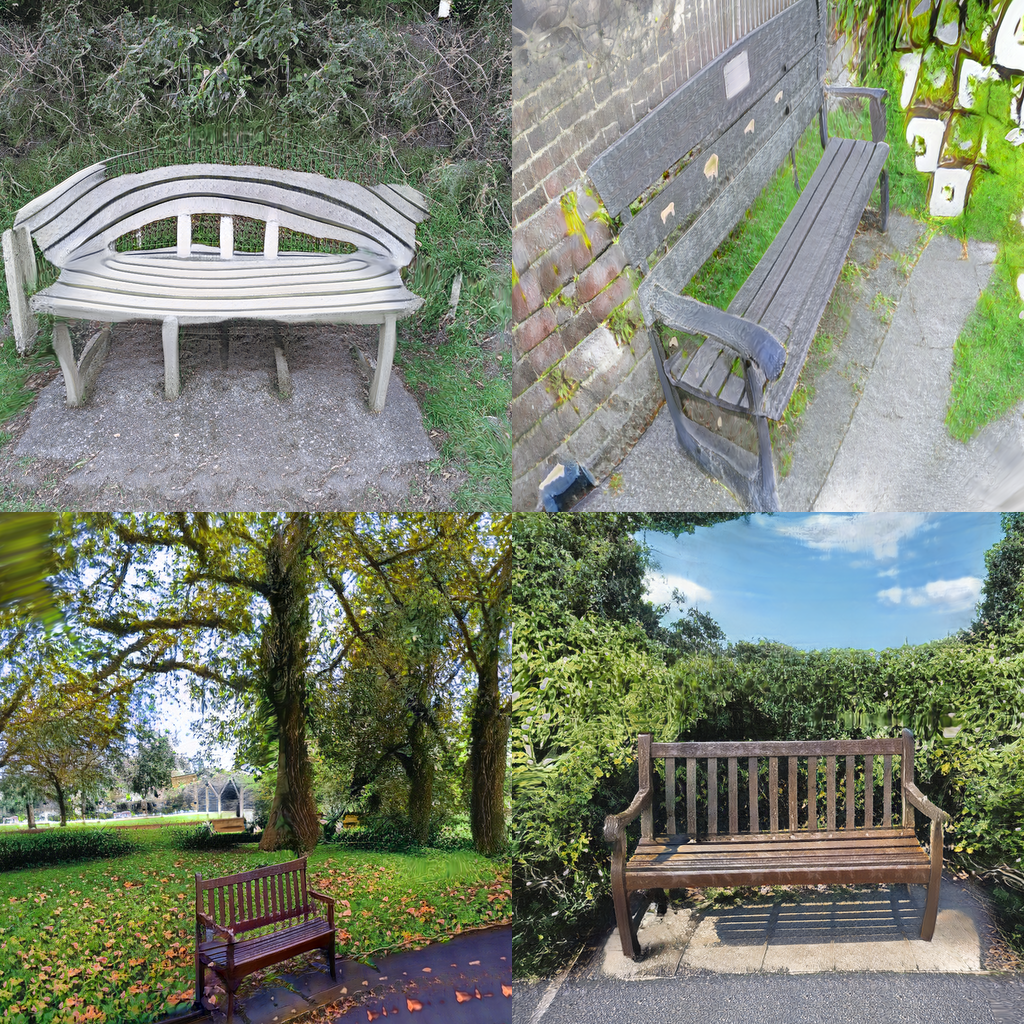}
    \caption{Benches}\label{subfig:Bench}
  \end{subfigure}\hfill
  \begin{subfigure}{0.19\linewidth}
    \includegraphics[width=\linewidth]{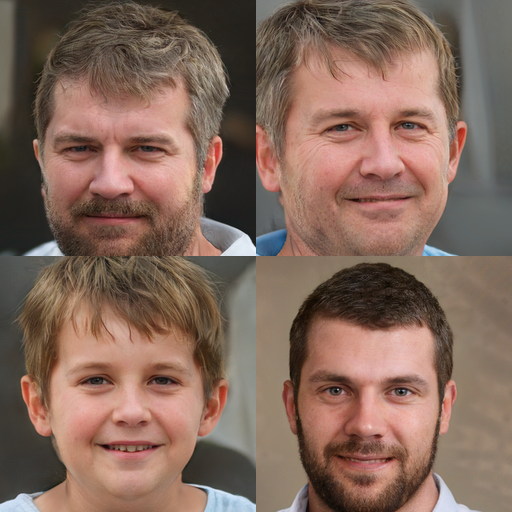}
    \caption{FFHQ-U}\label{subfig:FFHQ-U}
  \end{subfigure}\hfill
  \begin{subfigure}{0.19\linewidth}
    \includegraphics[width=\linewidth]{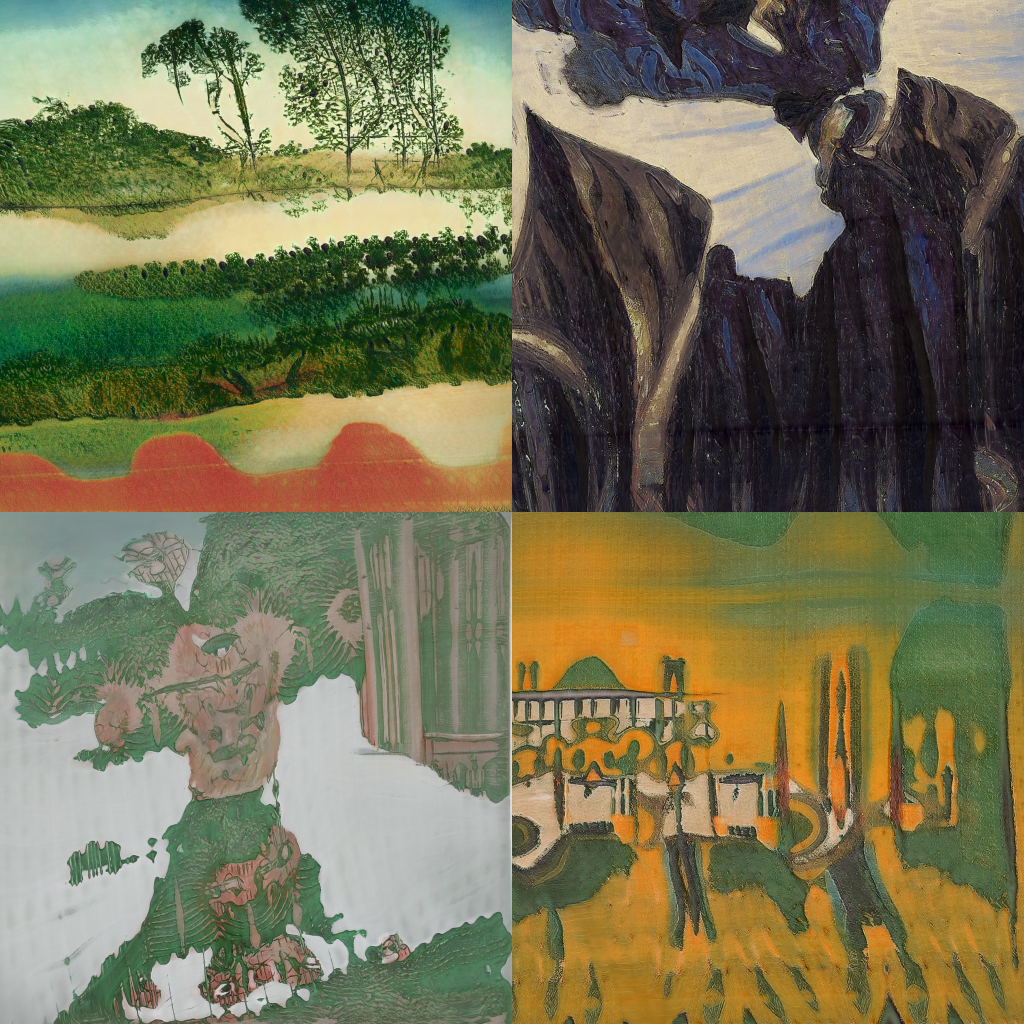}
    \caption{WikiArt}\label{subfig:WikiArt}
  \end{subfigure}\hfill
  \caption{(a): Four images (one image per class) on client \#1. (b), (c), (d), and (e): The images generated by different StyleGAN3s correspond to the aforementioned four classes.}
  \label{fig:generated}
\end{figure}

Although we adopt the pre-trained StyleGAN-XL by default as the server generator, our \method is also applicable to other StyleGANs due to the adaptable ability of our feature transformer ($F$). Here we consider utilizing the popular StyleGAN3~\cite{karras2021alias}, which has nearly $\frac{1}{3}$ of the parameter count compared to StyleGAN-XL. Specifically, we use several public StyleGAN3s pre-trained on four datasets with different resolutions: AFHQv2 ($512\times512$)~\cite{karras2021alias}, Benches ($512\times512$)~\cite{bowman2021trees}, FFHQ-U ($256\times256$)~\cite{karras2021alias}, and WikiArt ($1024\times1024$)~\cite{saleh2015large}. 
To adapt to different pre-trained generators, we re-tune the hyperparameter $\lambda$. According to \cref{tab:datasets_acc} and \cref{tab:generators}, our \method maintains excellent performance even when using other generators with different pre-training datasets. 
In \method, we prioritize the class-wise discrimination of the generated images over their semantic content. Thus, the knowledge-transfer-loop remains valuable when generated images are distinguishable by classes but do not share semantic relevance with clients' data (see \cref{fig:generated}).

\subsection{Iterative Domain Alignment Process}

\begin{figure}[ht]
  \begin{subfigure}{0.19\linewidth}
    \includegraphics[width=\linewidth]{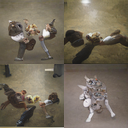}
    \caption{Iter. 0}\label{subfig:iter0}
  \end{subfigure}\hfill
  \begin{subfigure}{0.19\linewidth}
    \includegraphics[width=\linewidth]{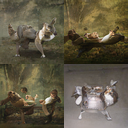}
    \caption{Iter. 1}\label{subfig:iter1}
  \end{subfigure}\hfill
  \begin{subfigure}{0.19\linewidth}
    \includegraphics[width=\linewidth]{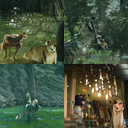}
    \caption{Iter. 10}
  \end{subfigure}\hfill
  \begin{subfigure}{0.19\linewidth}
    \includegraphics[width=\linewidth]{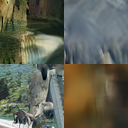}
    \caption{Iter. 20}
  \end{subfigure}\hfill
  \begin{subfigure}{0.19\linewidth}
    \includegraphics[width=\linewidth]{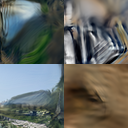}
    \caption{Iter. 30}
  \end{subfigure}
  \begin{subfigure}{0.19\linewidth}
    \includegraphics[width=\linewidth]{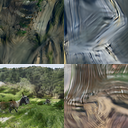}
    \caption{Iter. 50}
  \end{subfigure}\hfill
  \begin{subfigure}{0.19\linewidth}
    \includegraphics[width=\linewidth]{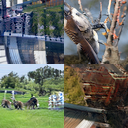}
    \caption{Iter. 100}
  \end{subfigure}\hfill
  \begin{subfigure}{0.19\linewidth}
    \includegraphics[width=\linewidth]{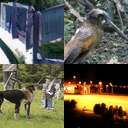}
    \caption{Iter. 110}
  \end{subfigure}\hfill
  \begin{subfigure}{0.19\linewidth}
    \includegraphics[width=\linewidth]{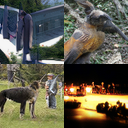}
    \caption{Iter. 120}
  \end{subfigure}\hfill
  \begin{subfigure}{0.19\linewidth}
    \includegraphics[width=\linewidth]{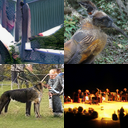}
    \caption{Iter. 130}
  \end{subfigure}
  \caption{The images generated by StyleGAN-XL correspond to four classes at different iterations.}
  \label{fig:align}
\end{figure}

The training process in HtFL is iterative, so the domain alignment in our \method is also an iterative process. Here we demonstrate the generated images throughout HtFL's training process in \cref{fig:align} to show the iterative domain alignment process. In the early iterations, as shown in \cref{subfig:iter0} and \cref{subfig:iter1}, the generated images ($\mathcal{D}_I$) corresponding to class-wise latent centroids ($\bar{\mathcal{Q}}$) appear similar, since clients cannot generate discriminative prototypes. As HtFL's training process continues, the generated images become increasingly class-discriminative and clear. The generated images in iterations 110, 120, and 130 hardly change for each class, showing the convergence of $F$ and client models' training. 

\subsection{Ablation Study}
\label{sec:ablation}

\begin{table}[ht]
  \centering
  \resizebox{\linewidth}{!}{
    \begin{tabular}{l|ccc|cc|c}
    \toprule
    \method & \textit{-$L_i^M$} & \textit{-$L^{\rm MSE}$} & \textit{-$L^{\rm MMD}$} & \textit{-ETF} & \textit{-$\bar{\mathcal{Q}}$} & \textit{+CS} \\
    \midrule
    28.17 & 24.39 & 21.70 & 20.14 & 21.02 & 20.69 & 24.13 \\
    \bottomrule
    \end{tabular}}
  \caption{The test accuracy (\%) of our \method's variants on Tiny-ImageNet in the practical setting using HtFE$_8$. }
    \label{tab:ablation}
    \vspace{-6pt}
\end{table}

\begin{figure}[ht]
  \begin{subfigure}{0.19\linewidth}
    \includegraphics[width=\linewidth]{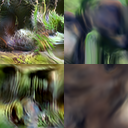}
    \caption{\textit{-$L_i^M$}}\label{subfig:lim}
  \end{subfigure}\hfill
  \begin{subfigure}{0.19\linewidth}
    \includegraphics[width=\linewidth]{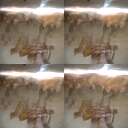}
    \caption{\textit{-$L^{\rm MSE}$}}\label{subfig:lmse}
  \end{subfigure}\hfill
  \begin{subfigure}{0.19\linewidth}
    \includegraphics[width=\linewidth]{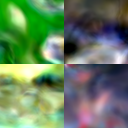}
    \caption{\textit{-$L^{\rm MMD}$}}\label{subfig:lmmd}
  \end{subfigure}\hfill
  \begin{subfigure}{0.19\linewidth}
    \includegraphics[width=\linewidth]{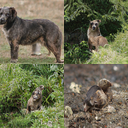}
    \caption{\textit{-ETF}}\label{subfig:etf}
  \end{subfigure}\hfill
  \begin{subfigure}{0.19\linewidth}
    \includegraphics[width=\linewidth]{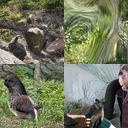}
    \caption{\textit{-$\bar{\mathcal{Q}}$}}\label{subfig:di}
  \end{subfigure}
  \caption{The images generated by StyleGAN-XL correspond to four classes in our \method's variants when variants converge. }
  \label{fig:ablation}
\end{figure}

\noindent Here, we remove $L_i^M$, $L^{\rm MMD}$, and $L^{\rm MSE}$ from \method and denote these variants ``\textit{-$L_i^M$}'', ``\textit{-$L^{\rm MMD}$}'', and ``\textit{-$L^{\rm MSE}$}'', respectively. 
Moreover, we create the following three variants. (1) ``\textit{-ETF}'': we remove $h_i$ and replace the ETF classifier with the original classifier of each model architecture. 
(2) ``\textit{-$\bar{\mathcal{Q}}$}'': we remove $L_i^M$ and mix the generated class-discriminative data $\mathcal{D}_I$ with local data $\mathcal{D}_i$. (3) Besides the common practice of using noise ${\bm \epsilon}$ to generate images, StyleGAN-XL offers a conditional version that can generate images belonging to any class from the ImageNet dataset. Using the Conditional StyleGAN-XL (CS), we create a variant ``\textit{+CS}'' by disabling step \textbf{\cir{2}} Upload and step \textbf{\cir{3}} Domain Alignment, and directly generating $C$ image-vector pairs for $C$ randomly selected ImageNet classes. 

The poor results of these variants in \cref{tab:ablation} and \cref{fig:ablation} demonstrate the effectiveness of each key component in our \method. Below, we analyze them one by one. 
(1) \textit{-$L_i^M$}: removing $L_i^M$ means training solely on the local dataset $\mathcal{D}_i$ without collaboration, leading to a \textbf{3.78\%} accuracy drop and distorted generated images (unused). 
(2) \textit{-$L^{\rm MSE}$}: 
removing $L^{\rm MSE}$ causes the generated images to become indiscriminative, thus misleading the local extractor and causing an accuracy drop of \textbf{6.47\%}. 
(3) \textit{-$L^{\rm MMD}$}: without the MMD loss for domain alignment, it is hard for $\bar{\mathcal{Q}}$ to be valid latent input vectors for the generator, leading to blurry images and a notable accuracy decrease. 
(4) \textit{-ETF}: biased classifiers make prototypes of different classes overlap, resulting in a loss of class-wise discrimination of the generated images. In \cref{subfig:etf}, three out of the four images depict dogs and grass. 
(5) \textit{-$\bar{\mathcal{Q}}$}: without $\bar{\mathcal{Q}}$, only using $\mathcal{D}_I$ on clients cannot transfer knowledge from the generator and mixing $\mathcal{D}_I$ and $\mathcal{D}_i$ perturb the semantics of local data, thus achieving poor performance and generating images with strange contents. 
(6) \textit{+CS}: using a conditional generator to produce class-wise image-vector pairs without adapting to clients' tasks can harm local training, as evidenced by a \textbf{0.26\%} decrease in accuracy compared to -$L_i^M$ (no collaboration). 
(7) Interestingly, the variants -$L^{\rm MSE}$, -$L^{\rm MMD}$, -ETF, and -$\bar{\mathcal{Q}}$ perform worse than -$L_i^M$, which indicates that all key components are crucial and assist each other in  \method.

\section{Conclusion}

We propose \method to promote client training in HtFL by (1) producing image-vector pairs that are related to clients' tasks through a pre-trained generator's inference on the server, and (2) transferring pre-existing knowledge from the generator to clients' heterogeneous models. Extensive experiments show the effectiveness, efficiency, and practicality of our \method in various scenarios.

\section*{Acknowledgements}

This work was supported by the National Key R\&D Program of China under Grant No.2022ZD0160504, the Program of Technology Innovation of the Science and Technology Commission of Shanghai Municipality (Granted No. 21511104700), and Tsinghua University(AIR)-Asiainfo Technologies (China) Inc. Joint Research Center. 

\appendix

We provide more details and results about our work in the appendices. Here are the contents:

\begin{itemize}
    \item \Cref{sec:sd}: The details and results of using the Stable Diffusion as the pre-trained generator on the server. 
    \item \Cref{sec:single}: The applicability of the knowledge transfer scheme of our \method in the scenario with only one server and one edge client. 
    \item \Cref{sec:addexp}: Additional experimental details, such as download web links of pre-trained generators, hyperparameter settings, \etc.
    \item \Cref{sec:privacy_cont}: The continued privacy-preserving discussion besides the main body with experimental results. 
    \item \Cref{sec:converge}: Empirical convergence analysis. 
    \item \Cref{sec:hyper}: The effects of using different hyperparameter settings for our \method. 
    \item \Cref{sec:add_abl}: Additional ablation study regarding the ArcFace loss. 
    \item \Cref{sec:visual}: Data distribution visualizations for different scenarios in our experiments. 
\end{itemize}

\section{Using the Stable Diffusion Model}
\label{sec:sd}

\subsection{Preliminaries}

\begin{figure}[h]
	\centering
	\includegraphics[width=\linewidth]{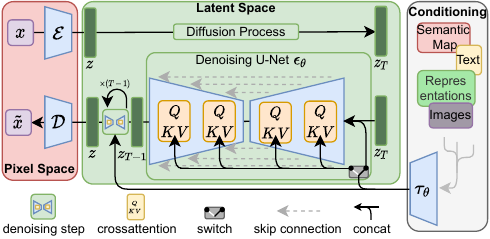}
	\caption{The main components in the Stable Diffusion~\cite{rombach2022high}.}
    \label{fig:sd}
\end{figure}

\noindent Most publicly available pre-trained generators can map from a latent vector (or a matrix, flattened to a vector) to an image, making them compatible with our \method. In Stable Diffusion~\cite{rombach2022high} (v1.5), the latent diffusion model (LDM) combined with VAE's decoder can map a latent vector $z_T$ to an image $\tilde{x}$, which is similar with StyleGANs~\cite{karras2019style, karras2020analyzing, karras2021alias, sauer2022StyleGAN} during generation, as shown in \cref{fig:sd}. Typically, $z_T$ is randomly generated from a normal distribution. Except for the above similarities, Stable Diffusion includes a conditioning component, which, for instance, can convert a text prompt to a conditional vector and influence the diffusion process to generate images with semantics related to the given prompt. As our \method is agnostic to the semantics of the images produced by the generator, one can select any valid text prompt, such as ``\textit{a cat},'' and maintain it unchanged throughout the entire FL process. 

\subsection{Experimental Results}

Due to the change in the pre-trained generator, we re-tuned some hyperparameters. Specifically, we set $\eta_s=0.1$, $\lambda=0.01$, and $\mu=100$ while maintaining the other hyperparameters consistent with those used for the StyleGAN-XL~\cite{sauer2022StyleGAN}. As shown in \cref{tab:sd}, using Stable Diffusion is also effective. While Stable Diffusion demonstrates excellent image generation performance, it's worth noting that the dimension of the latent vector is 16384, compared to 512 in the StyleGAN-XL. It is challenging to map low-dimensional client prototypes (with a dimension of 10 for the classification task on Cifar10) to such a high-dimension space while preserving their correlation. Perhaps a deeper feature transformer is required. 

\begin{figure*}[ht]
  \begin{subfigure}{\linewidth}
    \includegraphics[width=\linewidth]{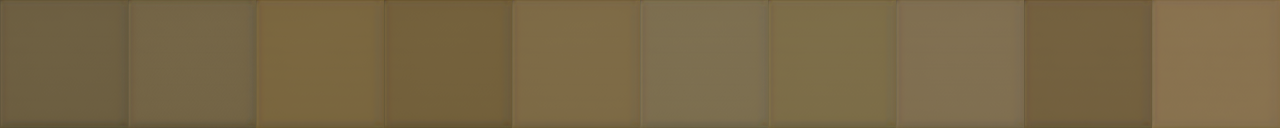}
    \caption{Iteration 0th}
  \end{subfigure}
  \begin{subfigure}{\linewidth}
    \includegraphics[width=\linewidth]{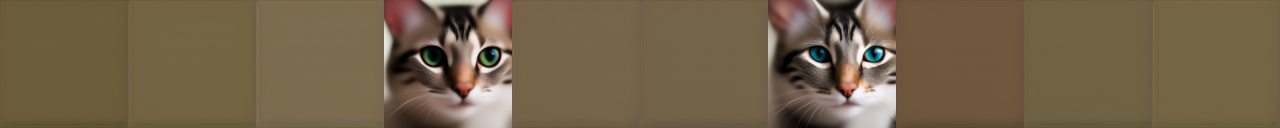}
    \caption{Iteration 20th}
  \end{subfigure}
  \begin{subfigure}{\linewidth}
    \includegraphics[width=\linewidth]{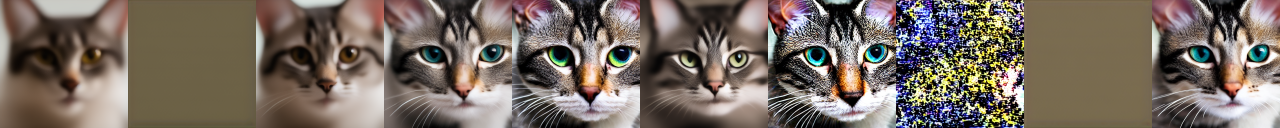}
    \caption{Iteration 40th}
  \end{subfigure}
  \begin{subfigure}{\linewidth}
    \includegraphics[width=\linewidth]{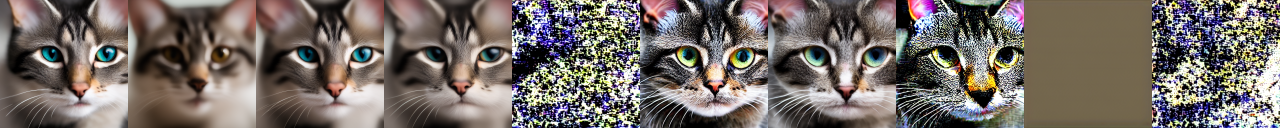}
    \caption{Iteration 60th}
  \end{subfigure}
  \begin{subfigure}{\linewidth}
    \includegraphics[width=\linewidth]{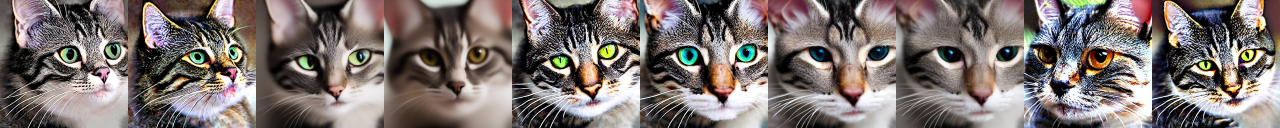}
    \caption{Iteration 80th} \label{80th}
  \end{subfigure}
  \caption{The prototypical images generated by Stable Diffusion corresponding to all 10 classes of Cifar10 at different communication iterations during the HtFL process.}
  \label{fig:sd_align}
\end{figure*}

We also show the generated images during the HtFL process in \cref{fig:sd_align}. With more iterations of HtFL, the generated images become clearer and more informative. Note that the label names in Cifar10 are ``airplane'', ``automobile'', ``bird'', ``cat'', ``deer'', ``dog'', ``frog'', ``horse'', ``ship'', and ``truck'' which correspond to labels from 0 to 9 and ten generated images in \cref{80th}. We observe that when label names have similar semantics compared to other labels, such as ``airplane'', ``automobile'', ``ship'', and ``truck'' (all being human-made vehicles), their corresponding generated images—like the 1st, 2nd, 9th, and 10th images in \cref{80th}—also exhibit similar characteristics, such as high resolution.

\begin{table}[ht]
  \centering
  \resizebox{!}{!}{
    \begin{tabular}{l|*{4}{c}}
    \toprule
    Generator & StyleGAN-XL & Stable Diffusion \\
    \midrule
    Accuracy & 87.63 & 87.71 \\
    \bottomrule
    \end{tabular}}
    \caption{The test accuracy (\%) of our \method with different pre-trained generators on Cifar10 in the practical setting using HtFE$_8$. }
    \label{tab:sd}
\end{table}

\section{The Scenario With a Single Edge Client}
\label{sec:single}

\begin{table*}[ht]
  \centering
  \resizebox{!}{!}{
    \begin{tabular}{l|*{4}{c}}
    \toprule
    Settings & 100-way 23-shot & 100-way 9-shot & 100-way 2-shot \\
    \midrule
    Client Data & 12.53$\pm$0.39 & 7.55$\pm$0.41 & 4.44$\pm$1.66 \\
    Our KTL & 13.02$\pm$0.43 & 8.88$\pm$0.62 & 8.76$\pm$2.25 \\
    \midrule
    Improvement & 0.49 & 1.33 & 4.32 \\
    Improvement Ratio & 3.91\% & 17.61\% & 97.29\% \\
    \bottomrule
    \end{tabular}}
    \caption{The test accuracy (\%) with Cifar100's subsets on a single client using a small model \ie, the 4-layer CNN. }
    \label{tab:single}
\end{table*}

\noindent In the traditional FL scenarios~\cite{mcmahan2017communication, zhang2022fedala}, clients mainly fetch extra knowledge from the globally aggregated model parameters. From the view of an individual client, these global model parameters contain fused knowledge from other clients. In our \method, except for the aggregated knowledge from clients, the pre-trained generator consists of common and valuable knowledge that can facilitate client training, particularly in addressing the data scarcity problem on edge devices. Therefore, the knowledge transfer scheme (\ie, the Knowledge-Transfer-Loop (KTL)) in our \method offers an additional feature \textit{beyond FL}, expanding its applicability to the scenarios with only one server and one edge client (\eg, the cloud-edge scenarios) and broadening the scope of its application. 

We can employ the KTL without modifying \method's workload, as the aggregation step has no effect with only one client. We iteratively execute the knowledge transfer process in each training epoch for this client until the training of the client model converges. Specifically, in each training epoch, the client sends a request (\ie, client prototypes) to the server, the server then sends a response (\ie, image-vector pairs) back to the client, and the response further serves as an additional supervised task to promote client training. 

By default, we adopt the StyleGAN-XL as the pre-trained generator on the server. On edge devices, data is usually insufficient~\cite{wu2020fedhome}. In our considered scenario, the edge client has a few training samples, which is the primary reason this client requires additional common knowledge. Specifically, we only assign $\frac{1}{20}$, $\frac{1}{50}$, and $\frac{1}{200}$ of the Cifar100 dataset to the client, respectively, where the number of samples is the same for all the 100 class. In other words, \textit{the client only has $23$, $9$, and $2$ training samples per class} in these three settings, respectively. 
From the view of few-shot learning~\cite{wang2020generalizing}, they are \textit{100-way 23-shot, 100-way 9-shot, and 100-way 2-shot} settings. 
Then, following the setting in the main body, we split the data into a training set (75\%) and a test set (25\%). 

As demonstrated in \cref{tab:single}, our KTL yields more improvement when the client has limited data, attributed to the introduction of additional pre-existing knowledge from the server-side pre-trained generator. However, according to the results for \textit{100-way 23-shot}, if the data on the edge client is not particularly scarce, simply introducing a pre-trained generator on the server without collaboration with other clients (such as using HtFL) can hardly bring improvement. These findings regarding KTL highlight its ability to transfer knowledge from a pre-trained model to an edge device with very limited data. 

\section{Additional Experimental Details}
\label{sec:addexp}

\noindent\textbf{Datasets, pre-trained generators, and environment. \ } 
We use four datasets with their respective download links: Cifar10\footnote{\url{https://pytorch.org/vision/main/generated/torchvision.datasets.CIFAR10.html}}, Cifar100\footnote{\url{https://pytorch.org/vision/stable/generated/torchvision.datasets.CIFAR100.html}}, Flowers102\footnote{\url{https://pytorch.org/vision/stable/generated/torchvision.datasets.Flowers102.html}}, and Tiny-ImageNet\footnote{\url{http://cs231n.stanford.edu/tiny-imagenet-200.zip}}. 
We can fetch the public pre-trained generators with their respective download links: StyleGAN-XL\footnote{\url{https://s3.eu-central-1.amazonaws.com/avg-projects/stylegan_xl/models/imagenet64.pkl}}, StyleGAN3 (pre-trained on AFHQv2)\footnote{\url{https://api.ngc.nvidia.com/v2/models/nvidia/research/stylegan3/versions/1/files/stylegan3-t-afhqv2-512x512.pkl}}, StyleGAN3 (pre-trained on Bench)\footnote{\url{https://g-75671f.f5dc97.75bc.dn.glob.us/benches/network-snapshot-011000.pkl}}, StyleGAN3 (pre-trained on FFHQ-U)\footnote{\url{https://api.ngc.nvidia.com/v2/models/nvidia/research/stylegan3/versions/1/files/stylegan3-r-ffhqu-256x256.pkl}}, StyleGAN3 (pre-trained on WikiArt)\footnote{\url{https://lambdalabs.com/blog/stylegan-3}}, and Stable Diffusion (v1.5)\footnote{\url{https://huggingface.co/runwayml/stable-diffusion-v1-5/tree/main}}. 
All our experiments are conducted on a machine with 64 Intel(R) Xeon(R) Platinum 8362 CPUs, 256G memory, eight NVIDIA 3090 GPUs, and Ubuntu 20.04.4 LTS. 

\noindent\textbf{Hyperparameter settings. \ } 
Besides the hyperparameter setting provided in the main body, we follow each baseline method's original paper for their respective hyperparameter settings. 
LG-FedAvg~\cite{liang2020think} has no additional hyperparameters. 
For FedGen~\cite{zhu2021data}, we set the noise dimension to 32, its generator learning rate to 0.1, its hidden dimension to be equal to the feature dimension, \ie, $K$, and its server learning epochs to 100. For FedGH~\cite{yi2023fedgh}, we set the server learning rate to be the same as the client learning rate, \ie, 0.01. 
For FML~\cite{shen2020federated}, we set its knowledge distillation hyperparameters $\alpha=0.5$ and $\beta=0.5$. For FedKD~\cite{wu2022communication}, we set its auxiliary model learning rate to be the same as the client learning rate, \ie, 0.01, $T_{start}=0.95$, and $T_{end}=0.95$. For FedDistill~\cite{jeong2018communication}, we set $\gamma=1$. For FedProto~\cite{tan2022fedproto}, we set $\lambda=0.1$. 
For our \method, we set $K=C$, $\mu=50$, $\lambda=1$, $\eta_S=0.01$ (server learning rate), $B_S=100$ (server batch size), and $E_S=100$ (the number of server training epochs), by using grid search in the following ranges on Tiny-ImageNet: 
\begin{itemize}
    \item $\mu$: $\{1, 10, 20, 50, 80, 100, 200\}$
    \item $\lambda$: $\{0.005, 0.01, 0.05, 0.1, 0.5, 1, 5, 10, 100\}$
    \item $\eta_S$: $\{0.0001, 0.001, 0.01, 0.1, 1\}$
    \item $B_S$: $\{1, 10, 50, 100, 200, 500\}$
    \item $E_S$: $\{1, 10, 50, 100, 200, 500, 1000\}$
\end{itemize}
Besides, we use Adam~\cite{kingma2015adam} for $F$ training following FedGen, set $s=64$ and $m=0.5$ following ArcFace loss~\cite{deng2019arcface}, and use the radial basis function (RBF) kernel for the kernel function $\kappa$ in $L^{\rm MMD}$. 
We use these settings for all the tasks. 

\noindent\textbf{Auxiliary model in FML and FedKD. \ } 
According to FedKD and FML, the auxiliary model needs to be designed as small as possible to reduce the communication overhead for model parameter transmitting, so we choose the smallest model to be the auxiliary model for FedKD and FML in any model heterogeneity scenarios. 

\section{Privacy-Preserving Discussion (Continued)}
\label{sec:privacy_cont}

Here we further discuss the privacy-preserving capability of our \method when a client has the potential to recover data from other clients. 
When a client receives additional \textit{\textbf{global knowledge}} (with data belonging to the labels never seen before), \textit{\textbf{the client is still unable to discern which image-vector pair belongs to which client (or group of clients), and thus cannot disclose the local data of other individual clients}}. As a result, transmitting class-level prototypes is a common practice in FL (\eg, FedProto~\cite{tan2022fedproto}). 
Secondly, in \textit{\S 3.3.5}, we have provided three reasons supporting the privacy-preserving capabilities of our \method based on its design philosophy. 
Moreover, our \method is \textit{\textbf{compatible with privacy-preserving techniques}}, such as adding noise, resulting in only a slight decrease in accuracy (see \Cref{tab:noise}). 

\begin{table}[ht]
  \centering
  \resizebox{!}{!}{
    \begin{tabular}{l|*{4}{c}}
    \toprule
     & --- & NC & NG & NC + NG \\
    \midrule
    \method & 53.16 & 52.73 & 51.16 & 50.51 \\
    \bottomrule
    \end{tabular}}
    \caption{The test accuracy (\%) on Flowers102 in the practical setting using HtFE$_8$ with \textit{noisy uploaded client prototypes} (NC) and \textit{noisy generated image-vector pairs} (NG). Following FedPCL~\cite{tan2022federated}'s privacy-preserving settings, we add Gaussian noise to the images and vectors before transmitting with a controllable parameters scale ($s$) and perturbation coefficient ($p$). We follow FedPCL to set $s=0.05$ and $p=0.2$ for vectors (or prototypes) and set $s=0.2$ and $p=0.2$ for images. }
    \label{tab:noise}
\end{table}

\begin{table*}[ht]
  \centering
  \resizebox{!}{!}{
    \begin{tabular}{l|ccc}
    \toprule
     & $K=C=102$ & $K=500$ & $K=1000$ \\
    \midrule
    Accuracy & 53.16 & 54.42 & 53.90 \\
    Upload & 0.07M & 0.35M & 0.69M \\
    \bottomrule
    \end{tabular}}
  \caption{The test accuracy (\%) and upload communication cost of our \method on Flowers102 in the practical setting using HtFE$_8$ with different $K$. ``M'' is shorter for a million. }
    \label{tab:K}
\end{table*}

\begin{table*}[ht]
  \centering
  \resizebox{!}{!}{
    \begin{tabular}{l|cccccc}
    \toprule
     & $\mu=1$ & $\mu=10$ & $\mu=20$ & $\mu=50$ & $\mu=100$ & $\mu=200$ \\
    \midrule
    Accuracy & 48.09 & 51.01 & 52.83 & 53.16 & 53.99 & 53.43 \\
    \bottomrule
    \end{tabular}}
  \caption{The test accuracy (\%) of our \method on Flowers102 in the practical setting using HtFE$_8$ with different $\mu$. }
    \label{tab:mu}
\end{table*}

\begin{table*}[ht]
  \centering
  \resizebox{!}{!}{
    \begin{tabular}{l|ccccc}
    \toprule
     & $\lambda=0.01$ & $\lambda=0.1$ & $\lambda=1$ & $\lambda=10$ & $\lambda=100$ \\
    \midrule
    Accuracy & 53.28 & 53.30 & 53.16 & 53.06 & 48.45 \\
    \bottomrule
    \end{tabular}}
  \caption{The test accuracy (\%) of our \method on Flowers102 in the practical setting using HtFE$_8$ with different $\lambda$. }
    \label{tab:lam}
\end{table*}

\begin{table*}[ht]
  \centering
  \resizebox{!}{!}{
    \begin{tabular}{l|cccc}
    \toprule
     & $\eta_S=0.0001$ & $\eta_S=0.001$ & $\eta_S=0.01$ & $\eta_S=0.1$ \\
    \midrule
    Accuracy & 49.84 & 51.51 & 53.16 & 53.62 \\
    \bottomrule
    \end{tabular}}
  \caption{The test accuracy (\%) of our \method on Flowers102 in the practical setting using HtFE$_8$ with different $\eta_S$. }
    \label{tab:eta_S}
\end{table*}

\begin{table*}[ht]
  \centering
  \resizebox{!}{!}{
    \begin{tabular}{l|ccccc}
    \toprule
     & $E_S=10$ & $E_S=50$ & $E_S=100$ & $E_S=200$ & $E_S=500$ \\
    \midrule
    Accuracy & 52.00 & 52.94 & 53.16 & 53.83 & 54.35 \\
    \bottomrule
    \end{tabular}}
  \caption{The test accuracy (\%) of our \method on Flowers102 in the practical setting using HtFE$_8$ with different $E_S$. }
    \label{tab:E_S}
\end{table*}

\begin{table*}[ht]
  \centering
  \resizebox{!}{!}{
    \begin{tabular}{l|ccccc}
    \toprule
     & $B_S=10$ & $B_S=50$ & $B_S=100$ & $B_S=200$ & $B_S=500$ \\
    \midrule
    Accuracy & 54.97 & 54.76 & 53.16 & 53.93 & 53.26 \\
    \bottomrule
    \end{tabular}}
  \caption{The test accuracy (\%) of our \method on Flowers102 in the practical setting using HtFE$_8$ with different $B_S$. }
    \label{tab:B_S}
\end{table*}

\section{Convergence Analysis}
\label{sec:converge}

\begin{figure}[h]
	\centering
	\includegraphics[width=\linewidth]{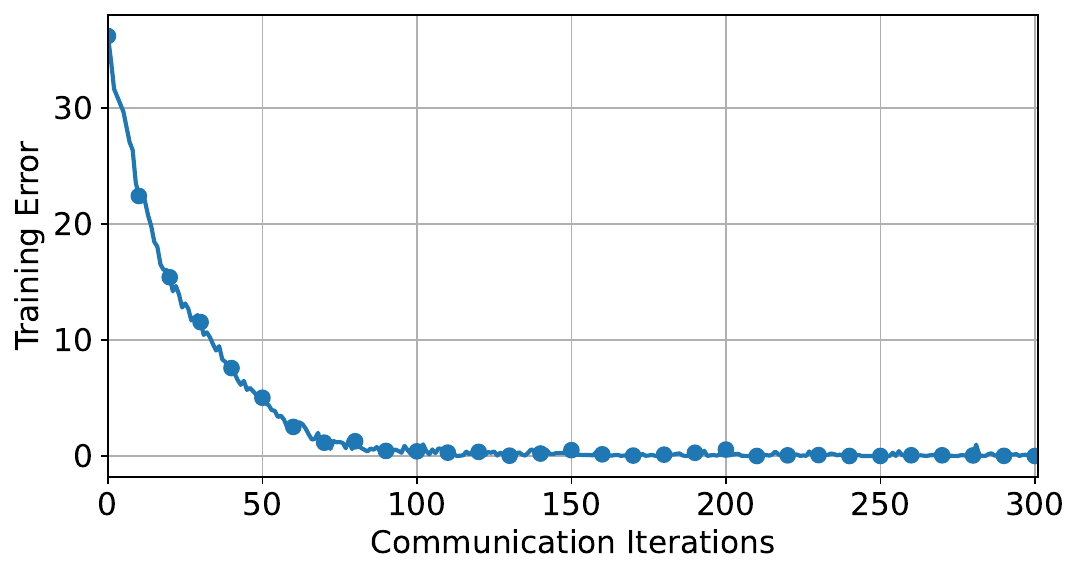}
	\caption{The training error curve for our \method on Flowers102 using HtFE$_8$ in the practical setting.}
    \label{fig:converge}
\end{figure}

We show the training error curve of our \method in \cref{fig:converge}, where we calculate the training error on clients' training sets in the same way as calculating test accuracy in the main body. According to \cref{fig:converge}, our \method optimizes quickly in the initial 80 iterations and gradually converges in the subsequent iterations. Besides, our \method maintains stable performance after converging at around the 120th iteration. 

\section{Hyperparameter Experiments}
\label{sec:hyper}

To study the effect of hyperparameters in our \method, we vary the value of each hyperparameter when keeping other parameters fixed, which are tuned on Tiny-ImageNet. Increasing the ETF dimension $K$ transmits more client knowledge to the server and improves accuracy, but this approach increases communication cost (see \cref{tab:K}). To save communication resources, we set $K=C$ in practice. According to \cref{tab:mu}, setting a value larger than 50 for $\mu$ can achieve an accuracy larger than 53\%, which means that the importance of $L_i^M$ should be emphasized. However, overly large values of $\mu$ can also lead to a decrease in accuracy. In contrast, in \cref{tab:lam} we find that the optimal value for the server's $\lambda$ is typically less than 10 on Flowers102, as too large values of $\lambda$ tend to weaken the domain alignment. Regarding the server hyperparameters $\eta_S$ and $E_S$, our \method can achieve better performance when using larger values for these parameters, as shown in \cref{tab:eta_S} and \cref{tab:E_S}. On the contrary, a smaller $B_S$ usually improves the performance of our \method (see \cref{tab:B_S}). 
In addition to the findings mentioned above, we also discover that the best combination of hyperparameters for Tiny-ImageNet is not necessarily the best for the Flowers102 dataset. While the default hyperparameter setting performs excellently, it is important to note that for a new dataset, one may need to re-tune the hyperparameters to achieve the best performance.

\section{Additional Ablation Study}
\label{sec:add_abl}

By default, in the main body, we adopt the ArcFace loss~\cite{deng2019arcface} as $L_i^A$. Specifically, we have
\begin{equation}
    L_i^A = \mathbb{E}_{({\bm x}, y) \sim \mathcal{D}_i} - \log{\frac{e^{s(\cos{(\theta_{y} + m)})}}{e^{s(\cos{(\theta_{y} + m)})} + \sum_{c=1, c \ne y}^C e^{s\cos{\theta_c}}}}, 
\end{equation}
where $\theta_{y}$ is the angle between $g_i({\bm x})$ and ${\bm v}_{y}$, $s$ and $m$ are the re-scale and additive hyperparameters~\cite{deng2019arcface}, respectively. Here, we adopt a more classical practice for $L_i^A$. Specifically, we replace the ArcFace loss with the contrastive loss~\cite{hayat2019gaussian, deng2019arcface}. In other words, we set $s=1$ and $m=0$ in $L_i^A$ to achieve this replacement, so we have
\begin{equation}
    L_i^A = \mathbb{E}_{({\bm x}, y) \sim \mathcal{D}_i} - \log{\frac{e^{\cos{\theta_{y}}}}{e^{\cos{\theta_{y}}} + \sum_{c=1, c \ne y}^C e^{\cos{\theta_c}}}}. \label{eq:cl}
\end{equation}
We denote this variant of \method as *$L_i^A$. 

\begin{table*}[ht]
  \centering
  \resizebox{!}{!}{
    \begin{tabular}{l|cccc}
    \toprule
     & Cifar10 & Cifar100 & Flowers102 & Tiny-ImageNet \\
    \midrule
    \method & 87.63 & 46.94 & 53.16 & 28.17 \\
    *$L_i^A$ & 85.28 & 41.63 & 51.30 & 27.52 \\
    \midrule
    $\Delta$ & -2.35 & -5.31 & -1.86 & -0.65 \\
    \bottomrule
    \end{tabular}}
  \caption{The test accuracy (\%) of our \method's variant *$L_i^A$ on four datasets in the practical setting using HtFE$_8$. }
    \label{tab:ablation1}
\end{table*}

We conduct experiments on four datasets using \cref{eq:cl} and show the test accuracy in \cref{tab:ablation1}. We observe that the impact of replacing $L_i^M$ varies across different datasets. However, it is consistent that removing the ArcFace loss leads to a decrease in accuracy. 

\section{Visualizations of Data Distributions}
\label{sec:visual}

We illustrate the data distributions (including training and test data) in the experiments here.

\begin{figure*}[ht]
	\centering
	\hfill
	\begin{subfigure}{0.47\linewidth}
	    \includegraphics[width=\textwidth]{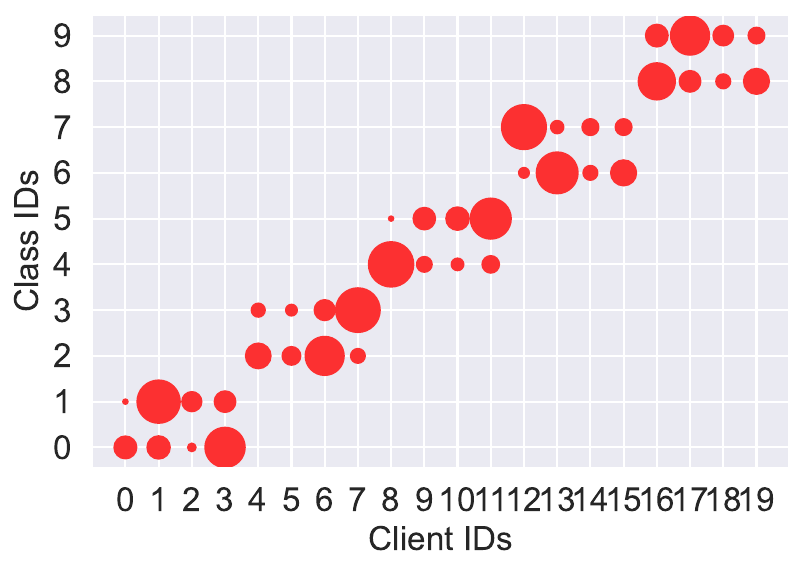}
	    \caption{Cifar10}
	\end{subfigure}
	\hfill
	\begin{subfigure}{0.49\linewidth}
	    \includegraphics[width=\textwidth]{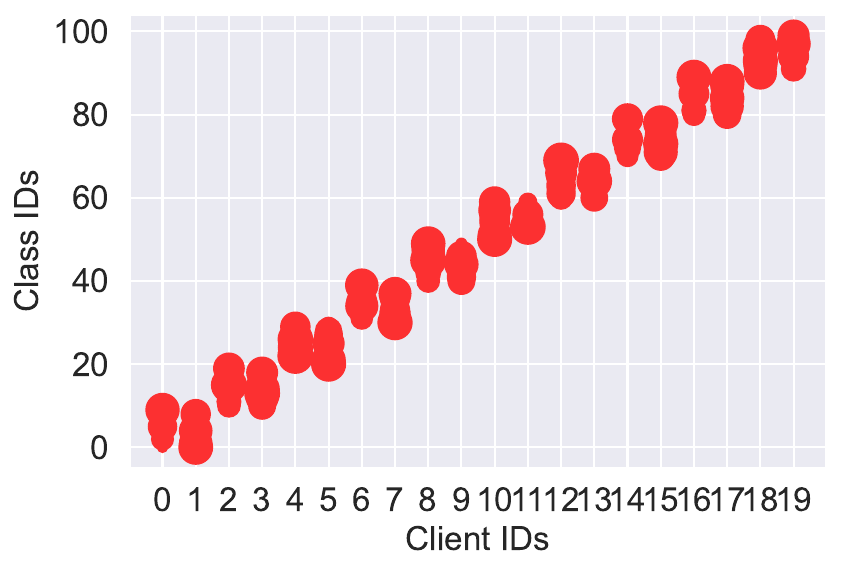}
	    \caption{Cifar100}
	\end{subfigure}
    \hfill
	\begin{subfigure}{0.48\linewidth}
	    \includegraphics[width=\textwidth]{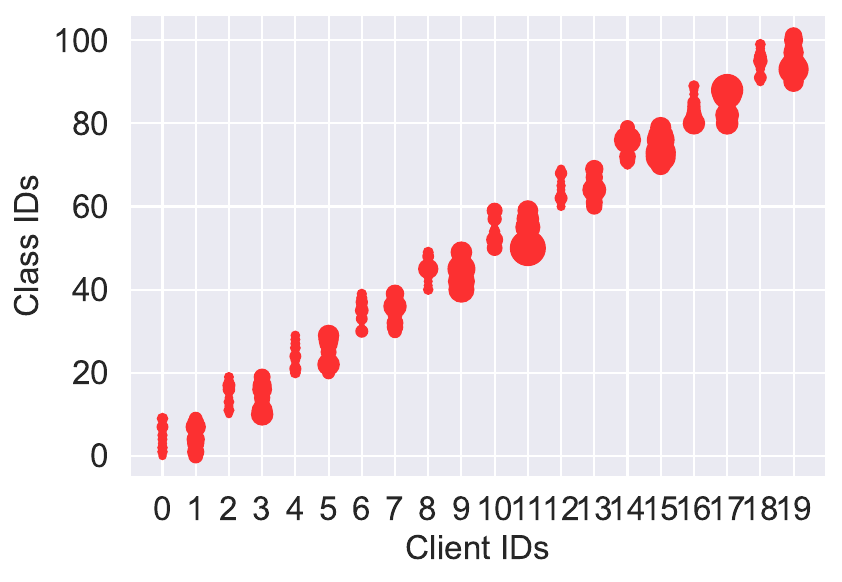}
	    \caption{Flowers102}
	\end{subfigure}
    \hfill
	\begin{subfigure}{0.48\linewidth}
	    \includegraphics[width=\textwidth]{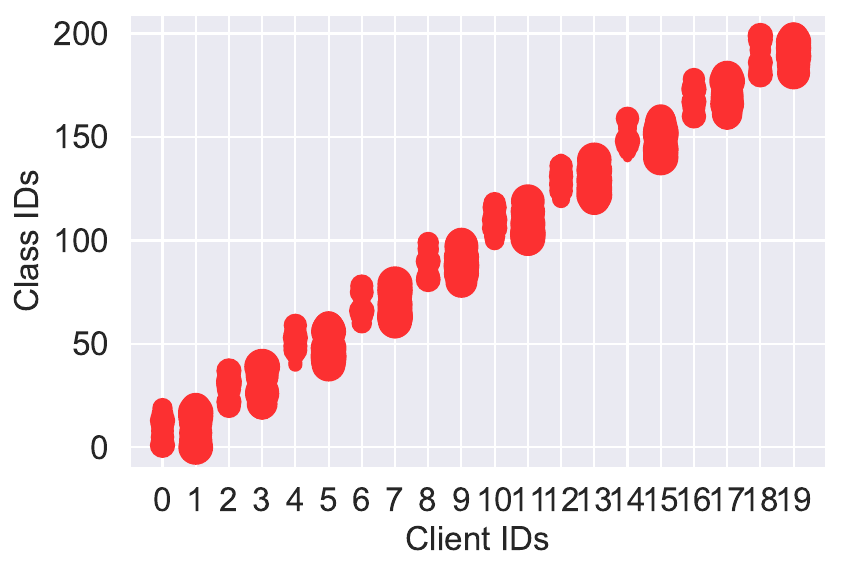}
	    \caption{Tiny-ImageNet}
	\end{subfigure}
    \hfill
	\caption{The data distribution of each client on Cifar10, Cifar100, Flowers102, and Tiny-ImageNet, respectively, in the pathological settings. The size of a circle represents the number of samples. }
	\label{fig:distribution-pathological}
\end{figure*}

\begin{figure*}[ht]
	\centering
    \hfill
	\begin{subfigure}{0.47\linewidth}
	    \includegraphics[width=\textwidth]{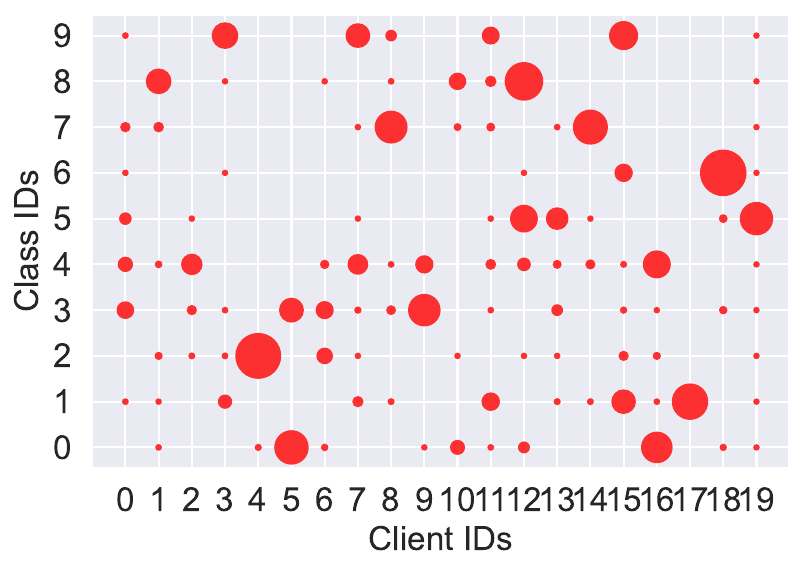}
	    \caption{Cifar10}
	\end{subfigure}
    \hfill
	\begin{subfigure}{0.49\linewidth}
	    \includegraphics[width=\textwidth]{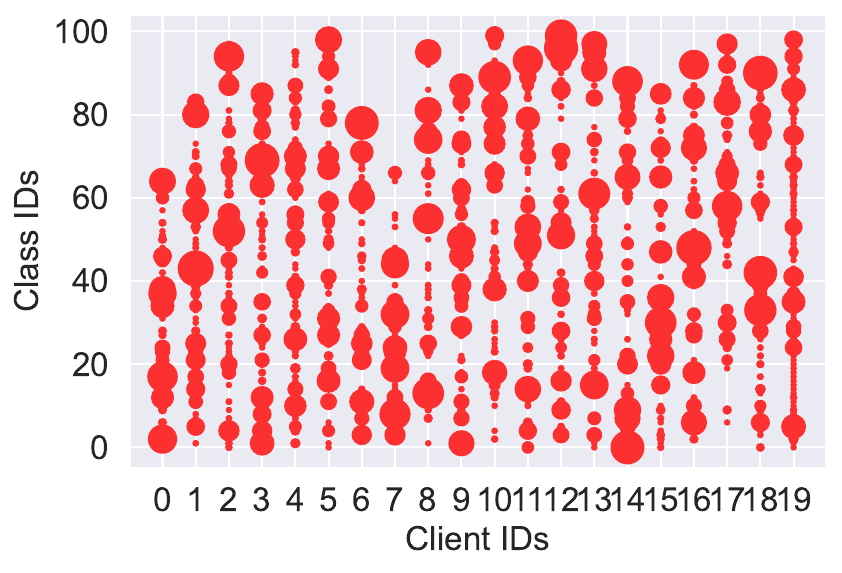}
	    \caption{Cifar100}
	\end{subfigure}
	\hfill
	\begin{subfigure}{0.48\linewidth}
	    \includegraphics[width=\textwidth]{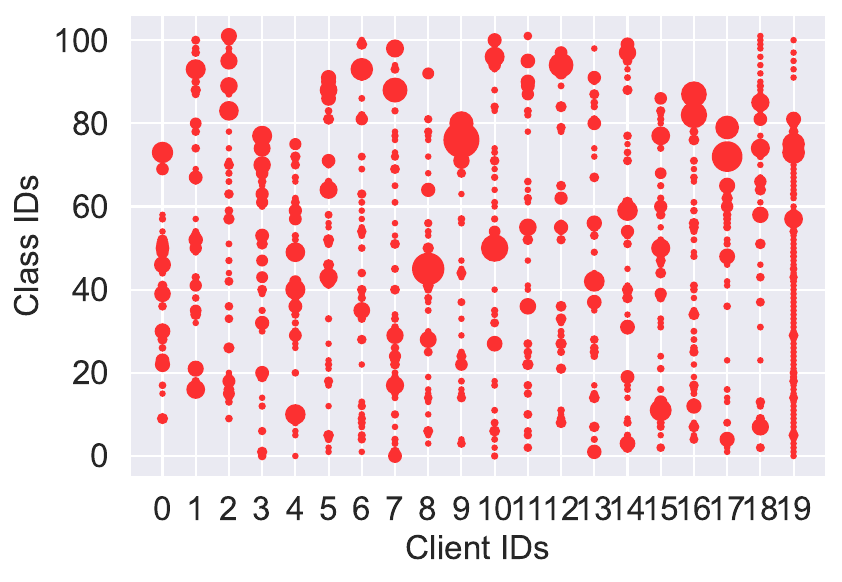}
	    \caption{Flowers102}
	\end{subfigure}
    \hfill
	\begin{subfigure}{0.48\linewidth}
	    \includegraphics[width=\textwidth]{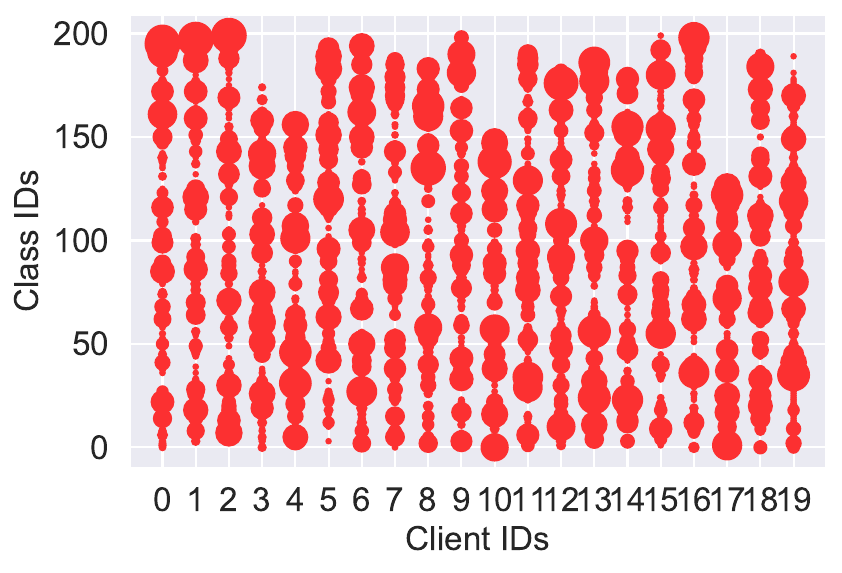}
	    \caption{Tiny-ImageNet}
	\end{subfigure}
    \hfill
	\caption{The data distribution of each client on Cifar10, Cifar100, Flowers102, and Tiny-ImageNet, respectively, in practical settings ($\beta=0.1$). The size of a circle represents the number of samples. }
	\label{fig:distribution-practical}
\end{figure*}

\begin{figure*}[ht]
	\centering
	\begin{subfigure}{\linewidth}
	    \includegraphics[width=\textwidth]{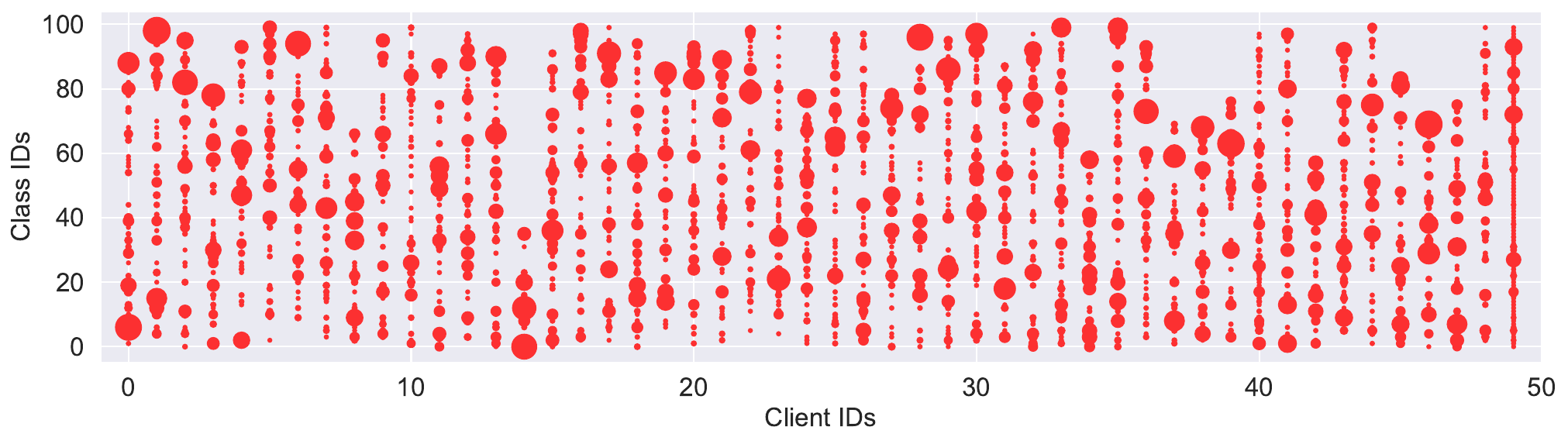}
	    \caption{50 clients}
	\end{subfigure}
	\hfill
	\begin{subfigure}{\linewidth}
	    \includegraphics[width=\textwidth]{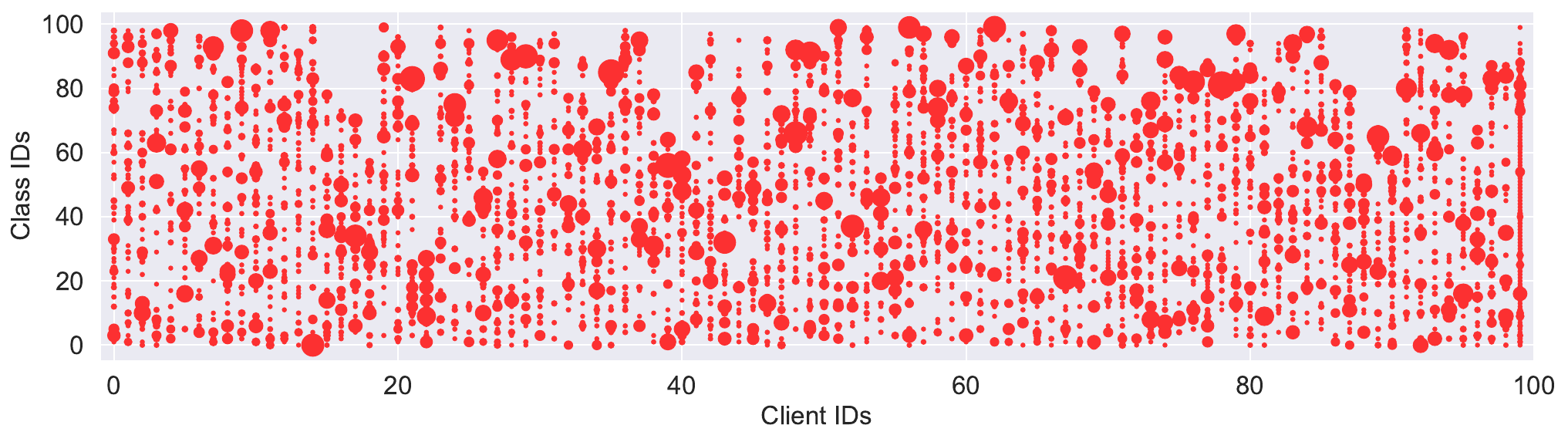}
	    \caption{100 clients}
	\end{subfigure}
	\hfill
	\begin{subfigure}{\linewidth}
	    \includegraphics[width=\textwidth]{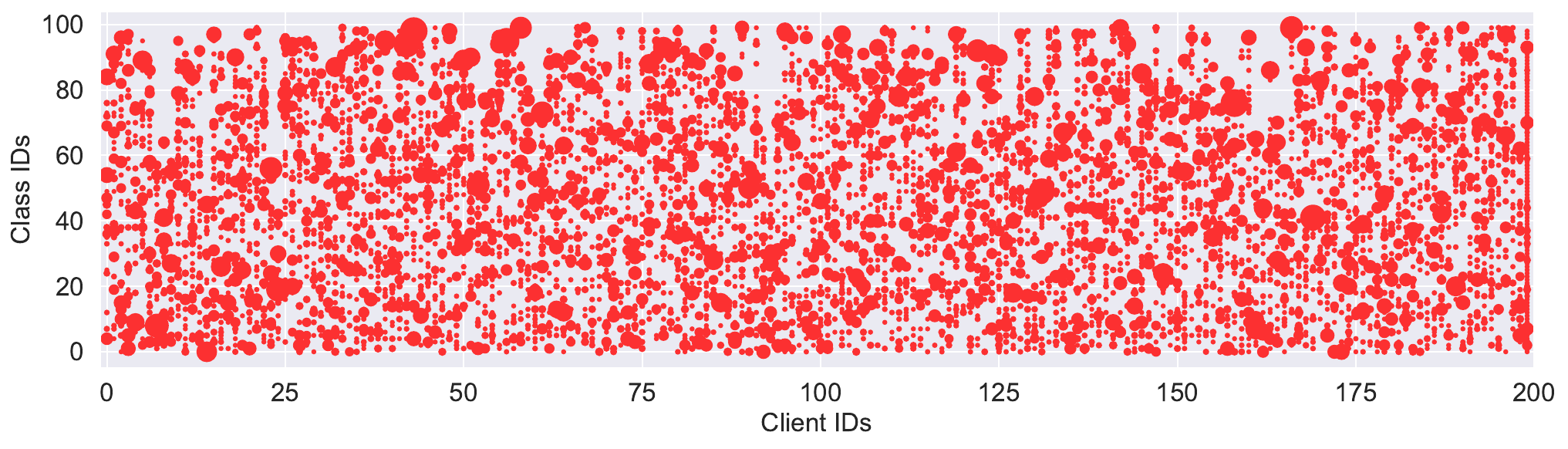}
	    \caption{200 clients}
	\end{subfigure}
	\hfill
	\caption{The data distribution of each client on Cifar100 in the practical setting ($\beta=0.1$) with 50, 100, and 200 clients, respectively. The size of a circle represents the number of samples. }
	\label{fig:distribution-50100}
\end{figure*}

\clearpage
{
    \small
    \bibliographystyle{ieeenat_fullname}
    \bibliography{main}
}

\end{document}

%% file: preamble.tex
\usepackage[dvipsnames]{xcolor}
